\documentclass{article}
\pdfoutput=1

\usepackage{microtype}
\usepackage{graphicx}
\usepackage{subcaption}
\usepackage{booktabs} % for professional tables

\usepackage{hyperref}
\usepackage{amsmath}
\usepackage{mathrsfs}
\usepackage{amsfonts}
\usepackage{amsthm}
\usepackage{arydshln}

\usepackage[arxiv]{icml2018mod}

\icmltitlerunning{Low-pass Recurrent Neural Networks}

\begin{document}

% We need  a different title
\twocolumn[
\icmltitle{Low-pass Recurrent Neural Networks -- A memory architecture for longer-term correlation discovery}

\icmlsetsymbol{equal}{*}

\begin{icmlauthorlist}
\icmlauthor{Thomas Stepleton}{dm}
\icmlauthor{Razvan Pascanu}{dm}
\icmlauthor{Will Dabney}{dm}
\icmlauthor{Siddhant M. Jayakumar}{dm}
\icmlauthor{Hubert Soyer}{dm}
\icmlauthor{Remi Munos}{dm}
\end{icmlauthorlist}

\icmlaffiliation{dm}{DeepMind Technologies, London, United Kingdom}

\icmlcorrespondingauthor{Thomas Stepleton}{stepleton@google.com}

\icmlkeywords{RNN, BPTT, timescale, low-pass}

\vskip 0.3in
]

\printAffiliationsAndNotice{}  

\begin{abstract}
Reinforcement learning (RL) agents performing complex tasks must be able to
remember observations and actions across sizable time intervals. This is
especially true during the initial learning stages, when exploratory behaviour
can increase the delay between specific actions and their effects. Many new or
popular approaches for learning these distant correlations employ
backpropagation through time (BPTT), but this technique requires storing
observation traces long enough to span the interval between cause and effect.
Besides memory demands, learning dynamics like vanishing gradients and
slow convergence due to infrequent weight updates can reduce BPTT's
practicality; meanwhile, although online recurrent network learning is
a developing topic, most approaches are not efficient enough to use as
replacements. We propose a simple, effective memory strategy that can extend
the window over which BPTT can learn without requiring longer traces. We
explore this approach empirically on a few tasks and discuss its implications.
\end{abstract}

\section{Introduction}

Numerous methods have been proposed to give neural networks the ability to
remember past observations. These range from LSTMs \citep{hochreiter1997long, Graves13Seq} and GRUs \citep{ChoMBB14} and extensions of
these \citep[e.g.][]{danihelka16} to ``external memory modules'' like DNC \cite{dnc} or Memory
Networks \cite{memnets}. Nearly all practical applications of these techniques rely
on backpropagation through time (BPTT) to compute weight updates. BPTT unrolls
the network over time, then applies backpropagation (reverse accumulation of
the chain rule), a computationally-efficient strategy with comparatively low
memory requirements.

Nevertheless, BPTT has drawbacks. The memory demand still increases
linearly with the size of the interval over which the network is unrolled,
as all observations and hidden activations throughout must be stored. This can be prohibitive beyond
a few hundred steps when observations are large. Partial mitigation strategies
and trade-offs do exist: \citep[e.g.][]{gruslys2016memory} uses careful bookkeeping to limit memory
usage at the cost of additional computation.

Beyond memory demands, BPTT applied to long sequences can encounter
vanishing and exploding gradients \cite{bengio1994learning, pascanu2013difficulty}. The algorithm recursively
multiplies the gradient over future steps by the Jacobian of the current
timestep, and the resulting products (which are summed) can be
unstable. Contractions to zero (``vanishing gradients'') or enormous
values (``exploding gradients'') are common problems. LSTM and GRU
architectures introduce gating mechanisms to mitigate vanishing gradients, and
constraining recurrent network weights can also help. Gradient clipping is widely used to address exploding gradients. Overall, though,
gradient misbehaviour can still be a difficult obstacle, particularly over long
intervals.

Finally, if a long trace of observations must be collected before BPTT
can compute a single weight update, infrequent gradient steps can slow
convergence significantly, (Indeed, we believe one of our experiments
demonstrates this.) In reinforcement learning, infrequent updates can
be especially problematic, since noisy credit assignment already makes
learning data-inefficient to begin with.

Given these issues, nearly all BPTT applications specifically use
\emph{truncated} BPTT, a variation that simply limits the network unroll
to a reasonably small fixed number of timesteps. Chunking the input data
in this way ensures manageable memory needs, frequent weight updates, and
fewer chances (i.e. matrix multiplications) for gradients to run amok.
Truncated BPTT obtains good results in many settings, including language
modeling \cite{ChelbaMSGBK13} and reinforcement learning \cite{mnih2016asynchronous}. However, if important
correlations do not occur within the same chunk, the network's ability to
learn them may be severely impaired.

In this work we explore a simple memory strategy that increases the effective
intervals over which correlations can be learnt well beyond the time windowing
inherent to truncated BPTT. We consider advantages and weaknesses of our
approach and highlight efficiency gains in several reinforcement learning
tasks.

\section{Recurrent Neural Networks with concrete}
\label{sec:lprnn}

Observations in naturalistic settings have rich underlying structure at
multiple timescales, from high-frequency details to slowly-changing contexts.
Although novel information arrives frequently, many features are recurrent
at time scales matching their permanence, like seasonal cycles (fixed)
or locomotive gaits (variable). Animals and humans have capabilities and habits
that exploit this structure. Grazing animals revisit seasonal foraging grounds,
where broad spatial recall aids efficient exploitation of known terrain whilst
omitting distracting detail. Humans also show varying precision for memories
of past events: we usually remember our distant past more coarsely than our
immediate past, and overcoming this structure tends to require kinds of
prosthesis (photographs, notes, etc.). This oversimplification omits
important memory capabilities: declarative memory, fine episodic recall of
events, and so on. Nevertheless, we suggest this ``logarithmic'' temporal
resolution forms desirable contexts for conditioning future behaviour,
a viewpoint echoing \cite{howard2018memory}.

\begin{figure}[t]
  \centering
  \includegraphics[width=\columnwidth]{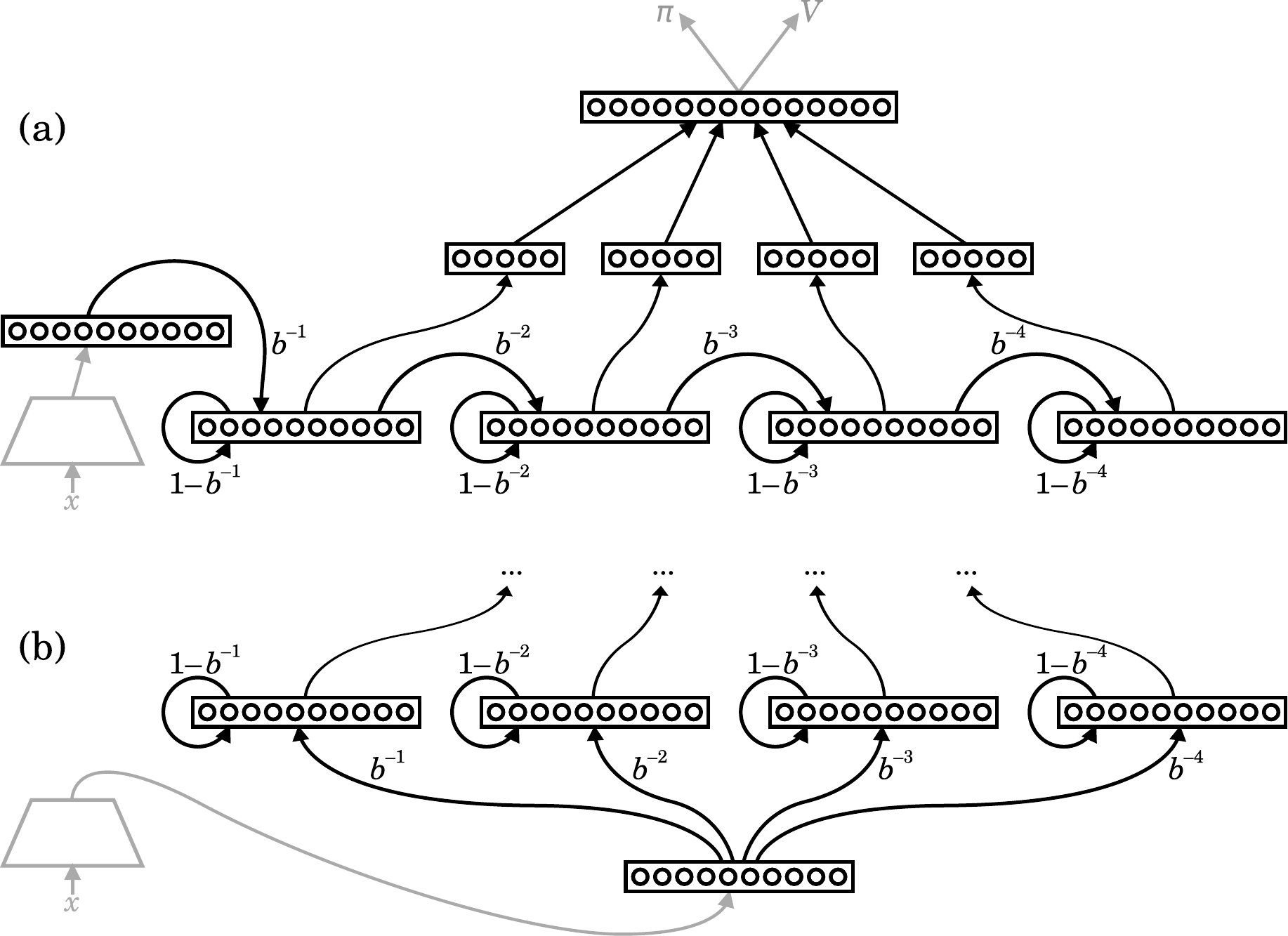}
  \caption{(a): Diagram of a Concrete memory system embedded within a
           simplification of the actor-critic network from the experiments.
           Parts in black are considered internal to the memory system by
           convention. An initial layer realises an input embedding (far
           left---here, the input is an observation $x$ processed by a
           convolutional network), which feeds into the chain of four
           low-pass-filtering pools forming the core of the memory. Chained
           pools have an exponentilly-diminishing smoothing factor with a
           fixed base $b$. Next, feedforward layers interpret pool contents:
           these comprise a set of per-pool layers (nicknamed {\em viewports})
           followed by a single unifying layer (a {\em summariser}). Subsequent
           network apparatus outside of the memory system yields behaviour and
           value judgments $\pi$ and $V$. In (b), a network used as an
           experimental control that replaces the chain with parallel filters.
           All components above the pools are identical to (a) and have been
           elided.}
  \label{fig:actcrit}
\end{figure}

Reflecting this structure, ``RNNs with concrete'', or more formally Low-Pass
Recurrent Neural Networks (LP-RNN) allocate varying representational precision
to events of different frequency/recency. They accomplish this via chains of
low-pass filtering pools, shown in their simplest form in
Figure~\ref{fig:actcrit}a, where each pool represents a different, broadening
portion of the network's past (Figure~\ref{fig:impulse}, also
Appendix~\ref{app:viz}). Our approach relates closely to other models that
exploit low-frequency content of inputs or activations, particularly
\cite{OlivaPS17,spears2017sith}, but our focus centres on our conceptually
simple model, its application to reinforcement learning, and its ability to
overcome limitations in correlation discovery imposed by truncated BPTT.

Formally, let us define the hidden state of some neural network as $\mathbf{h}_t= \mathtt{N}_{in}(\mathbf{o}_t)$, where $\mathbf{o}_t$ corresponds to the input of the model at step $t$, in this case some observation, and $\mathtt{N}_{in}$ is some neural network such as an MLP or convolutional network.

Then, in its simplest form, we specify LP-RNN as a chain of memory pools
(indexed by $n$) defined as: 
$$\mathbf{p}^{(n)}_t = a_n \mathbf{p}^{(n-1)}_t + (1-a_n) \mathbf{p}^{(n)}_{t-1},$$
where $\mathbf{p}^{(0)}_t = \mathbf{h}_t$ and $\mathbf{p}^{(n)}_0 = \mathbf{0}$,
and where optionally $a_n = b^{-n}$ for some base $b > 1$ to obtain a
regular temporal tiling as in Figure~\ref{fig:impulse}.
The output of the model can then be defined as some feedforward model $\mathtt{N}_{out}$, applied to the concatenation of all memory pools, that is $$\pi_t = \mathtt{N}_{out}([\mathbf{p}^{(0)}_t; \mathbf{p}^{(1)}_t;..; \mathbf{p}^{(k)}]).$$

This approach now allows a learning algorithm like BPTT to take advantage of
the historical information distributed in a windowed fashion across the memory
pools. In exchange for diminishing representational precision, an LP-RNN
accumulates data over far longer intervals than those over which the gradients
are backpropagated---even orders of magnitude longer.

For this reason, no off-the-shelf gradient-based method can alter the
slowly-changing representations in pools further down the chain---indeed, the
chained low-pass filters effectively extinguish gradients passing through, and
in practice we reduce compute needs by blocking gradients through all but the
``fastest'' pools. A network must learn to \emph{read} an LP-RNN---something a
feedforward network can do---and because gradients need not (or cannot) pass
beyond the reader, lengthy network unrolls have limited benefit. This is the
essence of LP-RNN's ``workaround'' for mitigating the limitations of BPTT.

(Of course, for architectures in Figure~\ref{fig:actcrit} and
Appendix~\ref{app:schem}, some gradients must pass through the LP-RNN to train
the input-processing apparatus to produce useful representations. For our tasks,
we find it suffices to pass gradients through the first, ``fastest'' pool only.
Different architectures and techniques like specifying auxiliary losses
\cite{jaderberg2017auxtasks,mirowski2017nav} are other avenues for ensuring that
data accumulating in the memory is rich and meaningful.)

LP-RNN's chained architecture stands in unique contrast to parallel multi-scale
filtering of the input (e.g. Figure~\ref{fig:actcrit}b, an architecture we
explore as an experimental control) or direct transformations of parallel
filters (see Related Work). We note that the arrangement presents appealing
opportunities for more intricate inter-pool links, such as random projections,
nonlinearities, or gating, which we hope to pursue in future.

For now, the ``odometer-like'', gradually-decelerating flow of data through an
LP-RNN, as well as its design for very long-term operation, recall to us Arthur
Ganson's kinetic sculpture \textit{Machine With Concrete}
\cite{ganson1992concrete}. As such, we have informally nicknamed our
model an ``RNN with concrete'', ``concrete memory'', or just ``Concrete'' for
short.

\section{Related Work}

LSTM and GRU, the workhorses of recurrent models, use gating to mitigate
vanishing gradients and (often) gradient clipping to limit divergence.
Most intuitive characterisations of these models present the gates as binary
``valves'' that either close to preserve stored data or open to replace it
with new input. In this account, LSTMs can retain information indefinitely; but,
as gates are actually computed as sigmoid functions applied to linear
projections of the input and prior hidden state, it is likely that they are
often neither fully open nor closed. Indeed, a binary-only behaviour would
disrupt gradient-based learning, which is not usually observed in practice.

Consider the formula for the LSTM memory cell update:
$$ \mathbf{m}_t = \mathbf{g}_{input} \circ \hat{\mathbf{h}}_t + 
\mathbf{g}_{forget} \circ \mathbf{m}_{t-1},$$
where $\mathbf{m}_{t}$ represents the state of the memory cell at time step $t$, $\mathbf{g}_{input}$ is the value of the input gate, $\mathbf{g}_{forget}$ the value of the forget gate, $\hat{\mathbf{h}}_t$ the proposed update for the memory, and $\circ$ element-wise multiplication. 
This form is reminiscent of a low pass filter; in fact, if we assume $\mathbf{g}_{forget} \approx 1 - \mathbf{g}_{input}$, then it is precisely a low pass filter. Tying the forget and input gate in this manner is used in practice \cite{greff2017lstm}; indeed, this low-pass update formula is hard-coded for GRU \cite{cho2014properties}.
We also observe that the earliest LSTM lacked forget gates altogether.

LSTM and GRU can thus be considered to filter their activations, but in
contrast to our work, the amount of filtering depends on the input and on
prior hidden-unit contents. Because it is learnt, the model can lose diversity
in filtering behaviour and focus on short term information early on in training.
If the forget gate operates independently of other gates, these models can also
reset themselves, giving them wavelet-like finite time support. Our simple model
fixes its filtering behaviour \emph{a priori}, and individual pools have
infinite (albeit temporally concentrated) support.

Another family of approaches augments recurrent models with external memory
storage. These approaches apply attention or nearest-neighbour lookups to
large buffers of activations, searching old hidden states to find associations
over long intervals. DNC and Memory Networks are canonical examples;
sequence-to-sequence models with attention \cite{bahdanau2014neural} also relate. In
reinforcement learning, episodic control methods \cite{blundell2016model, pritzel2017neural} do the same
to search over entire histories of agent observations. MBPA \cite{sprechmann2018memorybased} is another
relatable approach, used in supervised learning.

Some architectures address longer-term memory by storing information verbatim
over time intervals. NARX \cite{lin1996learning} derives new outputs from a
queue ("delay line")
of prior network inputs and outputs. More recently, Clockwork RNN
\cite{koutnik2014clockwork} has
units that only update at set intervals and otherwise hold their state. In
contrast to these fully-connected networks with non-lossy storage, LP-RNN
accepts information loss within a preset arrangement of pooled filtering units
specifically built to obtain the temporally-tiled impulse behaviour in 
Figure~\ref{fig:impulse}.

In \citet{Schmidhuber92a,ElHihi1995HRN,chung2017iclr} different recurrent multi-timescale architectures are explored, which are related to our work.  The architectures are hierarchical, where higher layer are slower, resulting in a different input response to ours. In \cite{chung2017iclr} the time-scale is learnt and is input dependent. \cite{Sordoni2015HRE, LingTDB15} exploits domain knowledge, providing input-driven time-scale boundaries for the different layers in the hierarchy. \cite{Fernandez2007SLS, KongDS15} have a similar objective to discover the segmentation structure by direct supervision.

More directly related works investigate various kinds of low-pass filtering
of inputs or activations. In reservoir computing, Echo State Networks benefit
considerably from ``leaky integration'' units: individual hidden units set up
to perform randomly-sampled amounts of low-pass filtering, where a
carefully-crafted distribution controls the range of available filters\cite{MantasLeaky}.
\cite{BandPassESN} extends this idea to to use band-pass instead of low-pass
filters. The serendipitous occurrence of an LP-RNN structure in a
randomly-constructed network is unlikely; even learning it in any
fully-connected architecture like those described above may be quite difficult.

Filtering has been applied to simpler recurrent networks with some minor
improvements \cite{BengioRNN13}. \cite{MikolovJCMR14} adds low-pass filtered versions of the network's
input to the inputs to the network's recurrent and output layers. The same
filtering is used throughout, so this approach is akin to having the first
pool in a Concrete memory project everywhere, provided the pool directly
filters the input and not some hidden state.

\cite{OlivaPS17} by contrast does present a recurrent model with multiple filterings
of hidden state, but these project via fully-connected recurrent connections
back into the model.
The fact that they are computed in parallel implies that models impulse response would be quite different from the Concrete architecture. The motivation of their approach is also quite different from ours. 
More recently, the neurologically-inspired SITH model \cite{spears2017sith}
proposes a somewhat more restricted variant of this concept that limits
projections of multiply-filtered \emph{network inputs} to those yielding
localised impulse responses similar to the ones in Figure~\ref{fig:impulse},
but this approach still employs full matrix multiplications to compute unit
values.

We would be pleased for Concrete to be considered a ``family member'' to the
above-mentioned works, but we consider the motivations, methods, applications,
and analysis in this paper---particularly the focus on overcoming limitations
of BPTT---to be unique.

Finally, there is recent work on online learning for recurrent models. Much
of it adapts Real Time Recurrent Learning
\cite{williams1989learning,pearlmutter1995gradient}, which applies forward
(instead of reverse) accumulation of the chain rule. Learning can be purely
online this way, but the method requires explicit computation and storage of
the Jacobian of the recurrent model's transition function. This can be
impracticable for large networks. \cite{tallec2017unbiased} explores various
low-rank Jacobian approximations. Taking a different approach, \cite{CzarneckiSJOVK17}
replaces BPTT with a learnt parameterised model of the gradient. These
approaches show promise, but none appear ready to replace truncated BPTT in
practical applications.

\section{Interpretive views on RNN with concrete}

\subsection{Signal processing perspective}

\begin{figure}[t]
  \begin{subfigure}{\columnwidth}
    \includegraphics[width=\columnwidth]{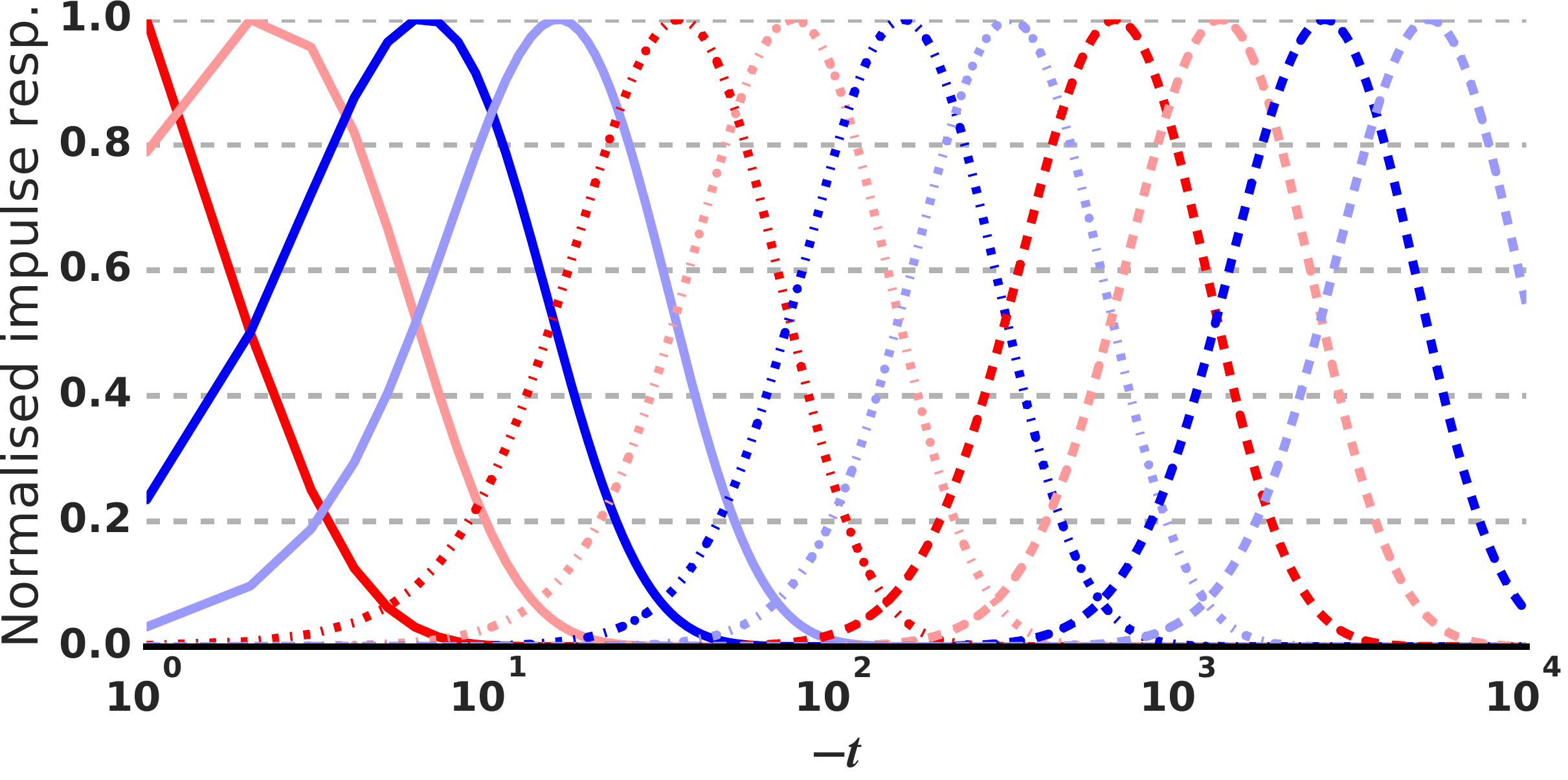}
    \caption{Normalised impulse responses of a chain of twelve Concrete pools;
             \textbf{note log scale}. Each curve shows the response of a single
             pool; the $y$-value is the degree to which an input at time $-t$
             in the past affects the contents of the pool, normalised by the
             maximum value for the pool. Sequential chaining of the pools causes
             the widening, temporally-tiled ``windows''.}
    \label{fig:impulse}

  \end{subfigure}
  \\[2mm]
  \begin{subfigure}{\columnwidth}
    \includegraphics[width=\columnwidth]{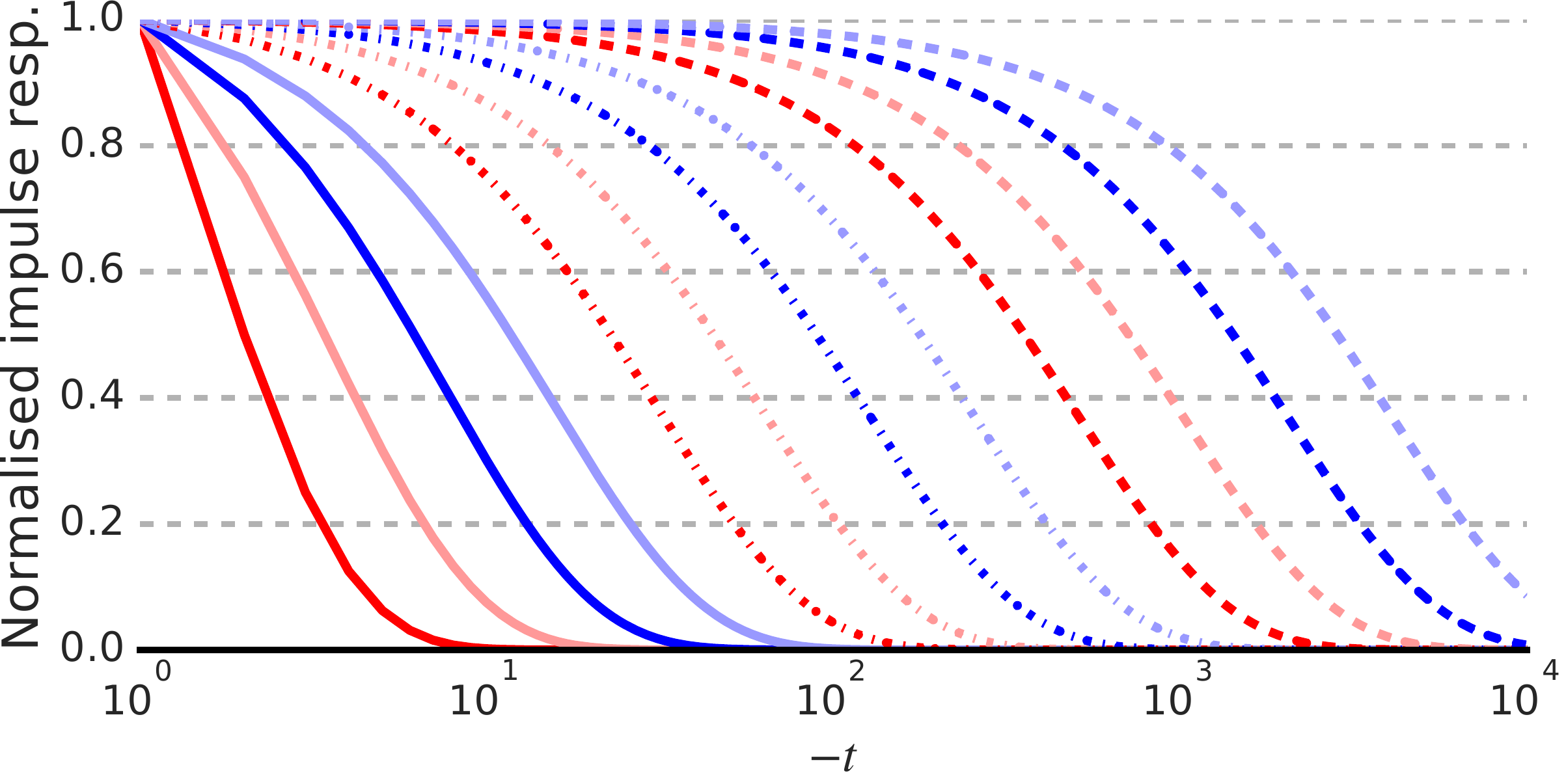}
    \caption{Contrasting with the above, normalised impulse responses of a
             a collection of twelve non-chained exponential smoothing pools
             (that is, the input to the memory is fed to each pool in parallel).
             Pools in this arrangement lack temporal tiling.}
    \label{fig:wrongcrete_impulse}
  \end{subfigure}
  \caption{Impulse responses for various filtering configurations. (Line styles
           and colours are only to make the pools more visually distinct.)}
\end{figure}

A natural first perspective views each memory pool as yielding a low-pass
filtered version of the input signal. This framing allows us to employ
classic signal processing methods to understand how the memory will behave
in practice.

For example, Figure~\ref{fig:impulse} shows the impulse response of the
different memory pools when they are chained, versus arranged in parallel in
Figure~\ref{fig:wrongcrete_impulse}. These curves obtain simply from feeding
unit values (an impulse) into the memory at the first timestep and zero values
thereafter. One observation is that the chained pools exhibit a behaviour
similar to a delay line, where information from the first pool moves to the
second and so on. Note that movement slows and temporal precision decreases
as we move through the pools ($x$-axis is log-scaled). 

The parallel pool case, where pools independently filter the input with
different smoothing factors, is quite different. Not only are we losing
resolution much faster over longer time spans, but we also lack the ``delay
line'' or ``temporal tiling'' effect. This is an important distinction, as an
MLP analysing parallel filter pools is less likely to localize information
temporally---not only due to the severe loss of resolution, but also because
recent events can affect the contents of all pools.

Another benefit of this perspective is insight into the kind of information the
memory is likely to retain: in particular, we can consider which frequency
bands are more or less likely to attenuate in the various pools. This allows us
to reason about what kind of cues the model is likely to remember at various
times (lower-frequency ones as time goes on).

\subsection{Linear Memories}

Consider the reinforcement learning setting, where the \textit{value} of a state $x$ is given as the sum of future discounted rewards, 
\begin{equation*}
    V(x) = \mathbb{E} \left[ \sum_{t=0}^{\infty} \gamma^t r_t | x_0 = x\right].
\end{equation*}
Observe that if the reward can be written as a linear function, $r_t = \phi(x_t)^\top w$, the value function decomposes as
\begin{equation*}
    V(x) = \mathbb{E} \left[ \sum_{t=0}^{\infty} \gamma^t \phi(x_t) | x_0 = x\right]^\top w = \psi(x)^\top w.
\end{equation*}
Here, $\psi(x)$ are the \textit{successor features} at $x$ for the current policy. Notice that if instead of a discounted return we want the value as the exponentially weighted moving average of rewards the same successor features can be used,
\begin{align*}
    \tilde V(x_t) &= (1 - \gamma) r_t + \gamma (1 - \gamma) r_{t+1} + \ldots.,\\
    &= (1 - \gamma) \psi(x_t)^\top w.
\end{align*}

Successor features \cite{dayan1993improving} provide a structural form of transfer learning in reinforcement learning. The features learned for one policy $\pi$ can be used to evaluate that policy under any reward function linear in $\phi$, with some corresponding $w$. This form of transfer owes to linearity. The reward function is assumed linear, and the prediction of interest here is itself a linear function of reward. Thus the same reward function can be applied to many predictions about the future (successor features), and vice versa.

Now consider predictions about the past: memories.

Let $f$ be a linear function on input, and $m$ be some memory system that maps a sequence of values $\{ f(x_0), \ldots, f(x_t) \}$ to a single value, $m(\{f(x_0)), \ldots, f(x_t) \})$. Different choices of $m$ would yield different time horizons on the memory, or different ``windows'' into the past.

If $m$ is linear in its inputs, we can write $m(\{f(x_t)\}) = f(m(\{x_t\}))$, leading to the same type of structural transfer learning seen by successor features. Specifically, any memory system $m$ can be applied to many different functions $f$, and vice versa. A simple example of how this could be useful is if we imagine a convolution over many memory systems, applying the same set of functions that ask the same questions about the past summarized in that memory. If the memory system is non-linear, such transfer is no longer possible. Each combination of $m \circ f$ must be learned in isolation, and particularly important, relearned if either changes. That is, if the receptive field in time of $m$ changes, but what $f$ computes about the state does not, both must be relearned if $m$ is non-linear, but only $m$ must change if it is linear.

Concrete is a linear memory system. Similarly, SITH is also a linear memory system as it combines several independent units using a linear operator. LSTMs by contrast are not linear memory systems.

\section{Experiments}
\subsection{Sequence classification tasks}
\label{sec:expsc}

\begin{figure}
  \includegraphics[width=\columnwidth]{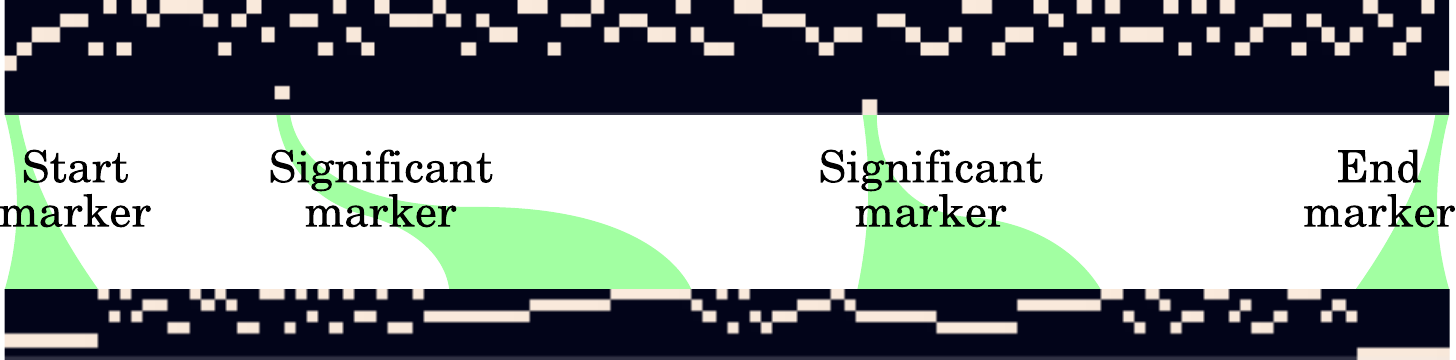}
  \caption{Visualisations of sequences of one-hot symbols for the original 
           four-class sequence classification task (top) and our modification
           with low-frequency symbol subsequences in place of single-timestep
           markers. The networks must classify the sequences based on the
           identity and ordering of these features. Not shown: the original task
           with three significant markers instead of two.}
  \label{fig:piano_roll}
\end{figure}

Our evaluation of the performance of our low-pass filtering-based memory
systems begins with the classic ``temporal order'' sequence classification tasks
from \cite{hochreiter1997long}. In these tasks, inputs are sequences of one-hot
symbols that start and end with unique boundary markers. In between, most
elements are any of four random distractor symbols, but two (or three) of the
elements can be either of two additional markers. (These elements appear in
specific regions only---e.g. the first within elements 10-20.) The
classification target of the sequence depends on which symbols appear in
which positions; so, if there are two (or three) significant markers, there
are $2^2$ (or $2^3$) possible classes. Sequences are between 100 and 110 symbols
long, and the training signal is supplied only at the end of the sequence.

In addition to the original tasks with two and three symbols, we also
investigate a variation of the two symbol task where the
semantically-significant binary markers are replaced by specific lower-frequency
subsequences of the distractor symbols. Each of the binary markers can be
replaced with any of five possible subsequences, none of which are shared
between the markers (hence, ten subsequences in total). Finally, the start and
end markers are also extended over multiple timesteps. These sequences can
be between 98 and 133 symbols long. See Figure \ref{fig:piano_roll} for
visualisations of both types of task.

For all tasks, we compare three families of classification network with memory:
an LSTM that projects into a hidden layer and then into a layer of unscaled
classification logits; a Concrete-like network without the input embedding found
in the actor-critic network in Figure \ref{fig:actcrit}a, and a network that
feeds an input embedding into a parallel (non-chained) bank of low-pass filters.
Detailed schematic diagrams of these network families appear in Appendix section
\ref{app:schem}. Our training system presents the network with a sequence of
temporally-contiguous, non-temporally-aligned minibatches, each compiled from
samples drawn from multiple independent sequence samplers running in parallel.
Because we vary the size of the minibatch and the amount of BPTT truncation,
we take care to ensure that the networks see the same number of input symbols
in each training instance ($4\!\times\!10^7$ for the original tasks; $10^8$ for
the modified task).

\begin{table}
  \begin{tabular}{llr}
  \toprule
  \textbf{Symbol}     & \textbf{Parameter}         & \textbf{Value set} \\
  \midrule
             & Batch size        & $\{4, 8, 16, 32, 64, 128\}$ \\
  $\alpha$   & Learning rate     & $[5\!\times\!10^{-7},\, 0.001]$ \\
  $\epsilon$ & Adam optimizer $\epsilon$ & $[5\!\times\!10^{-7},\, 0.001]$ \\
  $M$        & Hidden layer size & $\{16, 32, 64\}$ \\
  \hdashline
  $D$        & LSTM size         & $\{8, 16, 32, 64, 96\}$ \\
  $D$        & Pool size         & $\{8, 16, 24, 32, 48\}$ \\
  $k$        & Number of pools   & $\{4, 6, 8, 10, 12\}$ \\
  $V$        & Viewport size     & $\{4, 6, 10, 16\}$ \\
  $b$        & Pool decay base   & $\{1.5, 2. 3\}$ \\
  \bottomrule
  \end{tabular}
  \caption{Free parameters and the sets of values they are drawn from in the
           sequence classifcation tasks; notation matches the schematic diagrams
           in Appendix section \ref{app:schem}. Parameters above the dashed
           line are common to all network architectures; those below are
           sampled as appropriate for the network in use. The learning rate and
           epsilon parameters are sampled from a uniform distribution over the
           log of the values in the stated range.
           \textit{Adam optimizer} refers to~\cite{kingma2014adam}.}
  \label{tbl:scparams}
\end{table}

\begin{figure*}
  \begin{subfigure}{0.325\linewidth}
    \includegraphics[width=\linewidth]{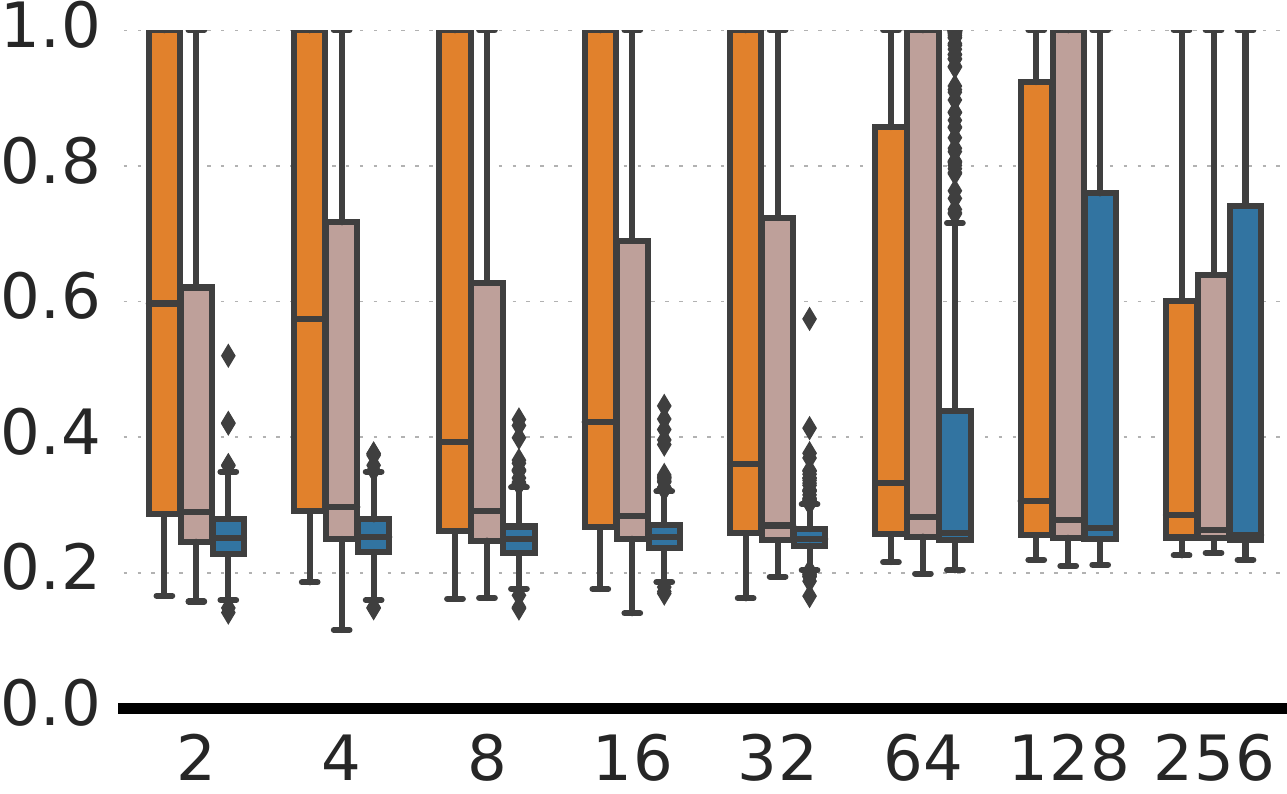}
    \caption{Two markers.}
    \label{fig:sca}
  \end{subfigure}
  \hspace*{\fill}
  \begin{subfigure}{0.325\linewidth}
    \includegraphics[width=\linewidth]{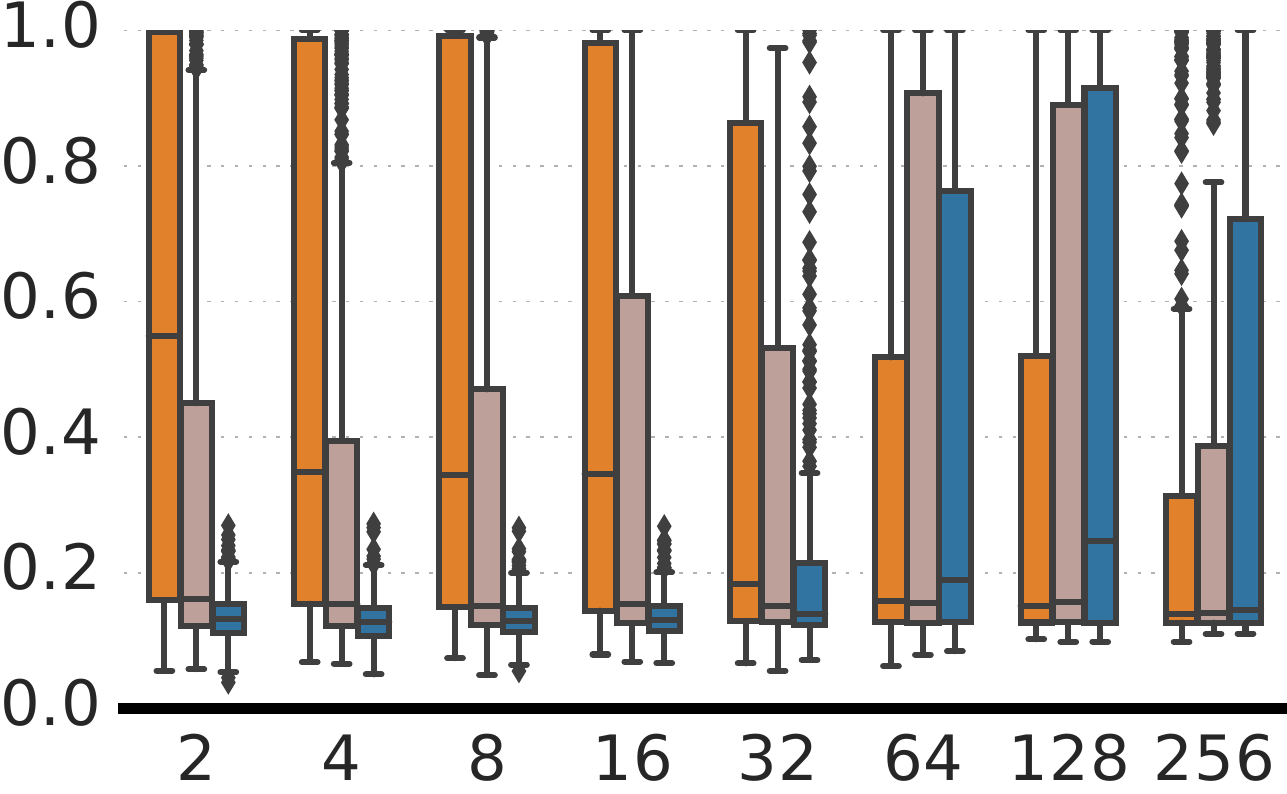}
    \caption{Three markers.}
    \label{fig:scb}
  \end{subfigure}
  \hspace*{\fill}
  \begin{subfigure}{0.325\linewidth}
    \includegraphics[width=\linewidth]{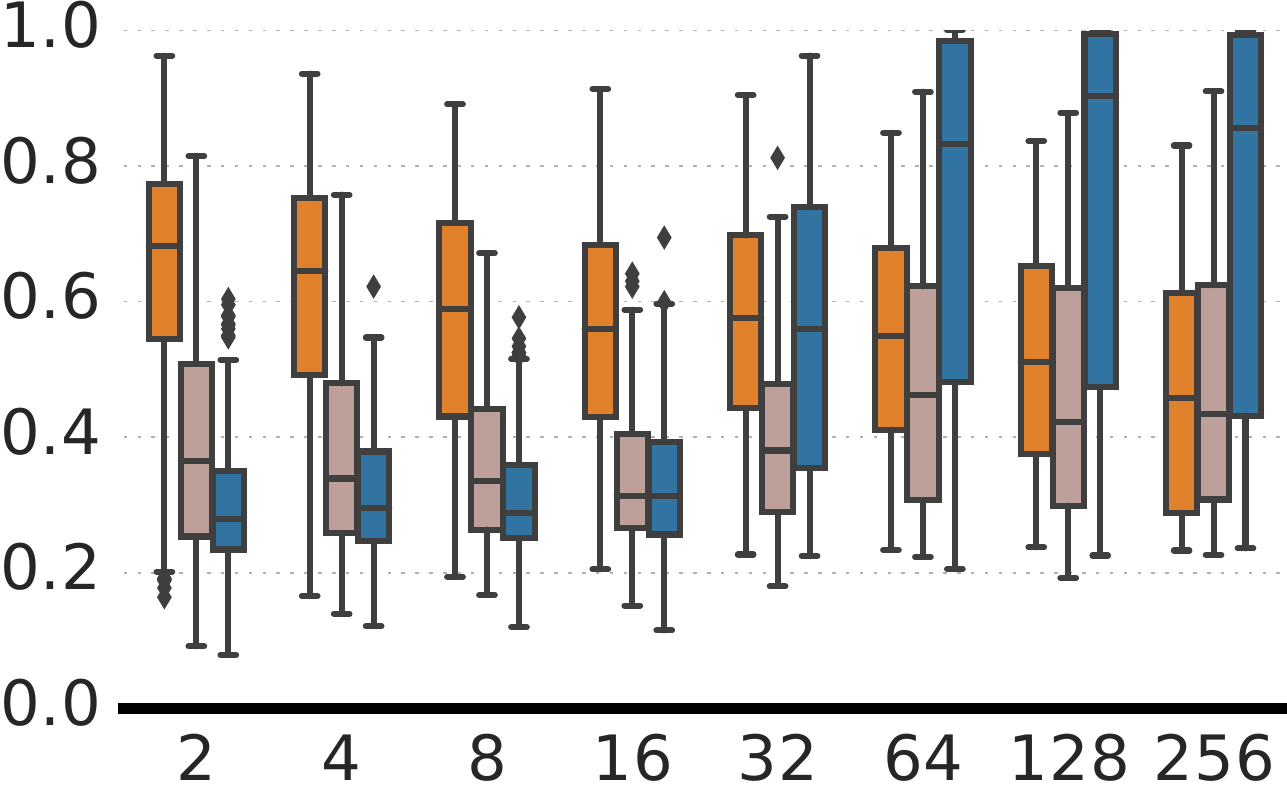}
    \caption{Two subsequence markers.}
    \label{fig:scc}
  \end{subfigure} \\[2mm]
  \includegraphics[width=0.45\linewidth]{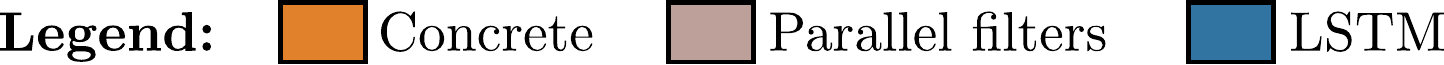}

  \caption{Performance on ``sequence classification'' tasks for networks with
           Concrete-style, parallel filter-based, and LSTM memory
           systems, broken down by BPTT truncation length ($x$ axis). The $y$
           axis is the fraction of correct classifications at the end of
           training as determined by averaging recent runs with an exponential
           smoother (smoothing factor: 0.98). For all tasks, the filter-based
           memory systems show better top performance than LSTM until the
           truncation length approaches the length of the input sequences.
           Each condition aggregates over a wide range of free hyperparameters
           (see Table \ref{tbl:scparams}), so some poor performance is not
           necessarily surprising; more important is that some instances of the
           LSTM-based network have the capacity to master the tasks but cannot
           do so under aggressive truncation. See Appendix section
           \ref{app:suppsc} for further discussion of performance trends.}
  \label{fig:sc}
\end{figure*}

For each task and network type, we perform 3,000 independent experimental runs
with various network and training hyperparameters drawn randomly from the sets
shown in Table \ref{tbl:scparams}. These sets express rather loose bounds on
favourable parameters (with the parameter space for the low-pass
networks being considerably larger than that for LSTM) and include
configurations for all memories where the hypothetical maximum capacity is
well in excess of what the tasks require. Put differently, all tasks are
given numerous architectural opportunities to succeed. Nevertheless, Figure
\ref{fig:sc} demonstrates that the LSTM-based network cannot solve the tasks
under any configuration until BPTT truncation covers a long-enough time
interval. (Low-pass filter-based memories do show a performance drop as
truncation relaxes, but we attribute at least some of this to sensitivity
to the volume of data and not its contents, as a similar effect occurs for
minibatch size. See Appendix section \ref{app:suppsc}.)

Finally, we note better performance for the Concrete-style memory over the
parallel filter arrangement, an effect that may owe to Concrete's more
temporally-localised response characteristics (Figure \ref{fig:impulse}).

\subsection{Reinforcement learning tasks}
\label{sec:exprl}

Reinforcement learning is a motivating application for our investigation into
memory systems based on low-pass filtering. The training signal in RL settings
can be sporadic and noisy, making it difficult to identify the information
that may be relevant to future behaviour. Furthermore, some settings feature
long time intervals between the moments when information is acquired and when
it must be acted upon. We examine three tasks that exacerbate this latter
challenge: in each, the network must retain relevant information for hundreds
of timesteps.

For all three tasks, we compare an actor-critic network that uses a
Concrete-style memory with a network that uses an LSTM. The networks are
identical outside of their memory systems: both feature the same convolutional
network and hidden layer to process the input; both have the same linear
outputs for computing state values and logits over discrete actions. (Detailed
schematics of the networks in use appear in Appendix section \ref{app:schem}.)
We use the IMPALA distributed reinforcement learning architecture
\cite{espeholt2018impala} with an adaptation of Population Based Training (PBT) \cite{jaderberg2017pbt}. PBT is a recently proposed meta-algorithm that optimises hyperparameters and network weights through time among a population
of agents.
In our adaptation, agents copy the weights of better-performing agents whilst
keeping their own hyperparameters fixed, effecting a parallelised exploration
of weight space that exposes promising weight configurations to varied
hyperparameter regimes.
Further implementation details are given in the appendix \ref{app:pbt}.
Because IMPALA uses a multi-step value learning
algorithm for the critic, rollout length affects RL efficiency in addition
to BPTT truncation; to avoid conflating these effects, we fix all rollouts at
300 timesteps and artificially block gradients through the network's hidden
state at varying intervals.

All three tasks are ``gridworld'' environments implemented with the pycolab game
engine \cite{tomstepleton2017}. Although recent results in RL emphasise more
elaborate settings \cite{beattie2016deepmind, brockman2016openai} our own
experience suggests that the intricacies of these games can give the agent
opportunities to ``cheat'' in memory tasks by coding information in the
environment itself. An agent may keep to one side of a 3-D corridor, for
example, to remember which direction to turn at an upcoming junction. Our
custom environments allow us to control precisely what the agent can do in
the environment, thereby letting us force it to store critical information in
its memory.\footnote{Implementations of these environments may be found at
\url{https://github.com/deepmind/pycolab/tree/master/pycolab/examples/research/lp-rnn}.}

\subsubsection{Cued Catch}

\begin{figure*}
  \begin{subfigure}{0.24\linewidth}
    \includegraphics[width=\linewidth]{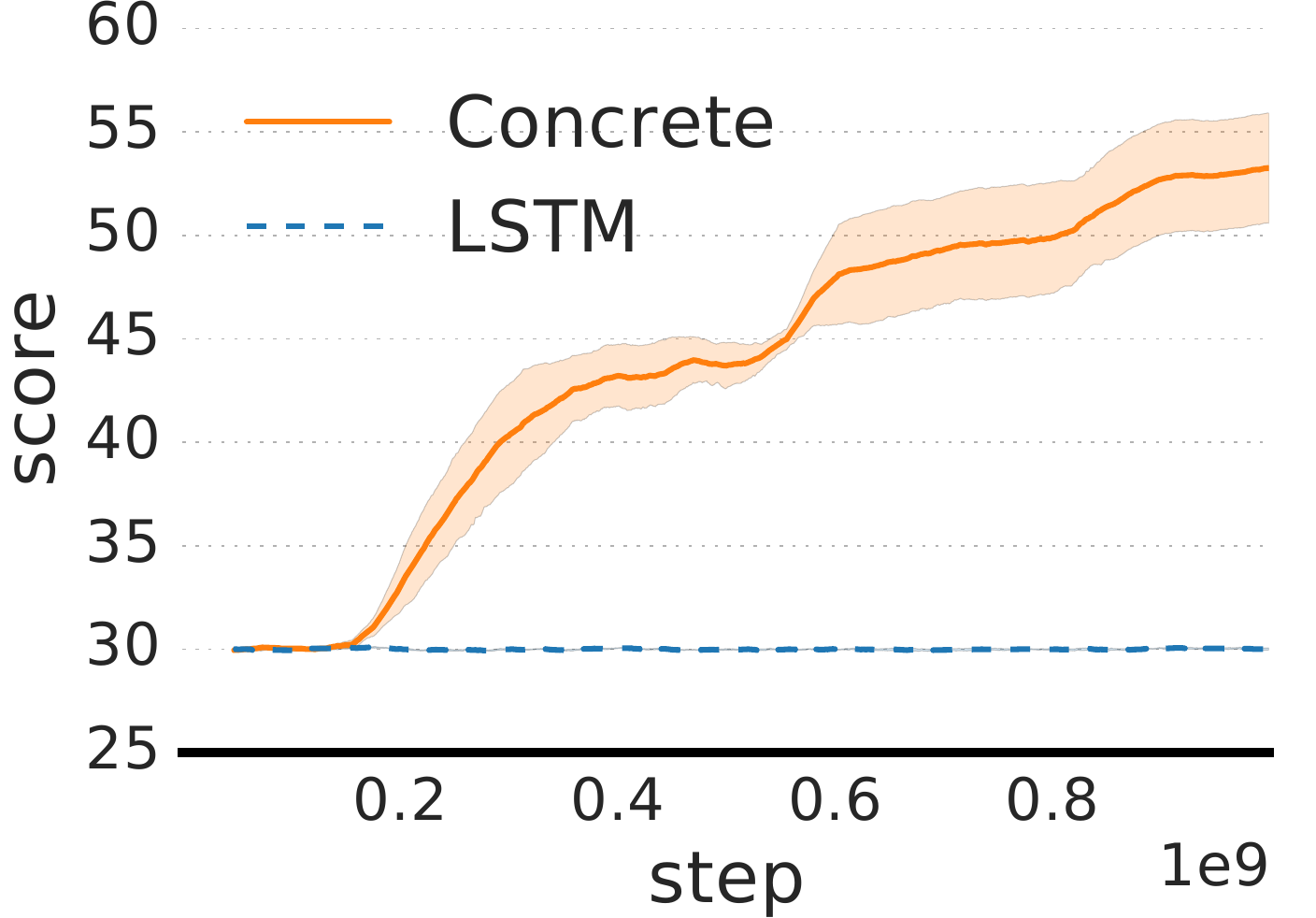}
    \caption*{\em\hspace{4mm}Every 3 steps.}
  \end{subfigure}
  \hspace*{\fill}
  \begin{subfigure}{0.24\linewidth}
    \includegraphics[width=\linewidth]{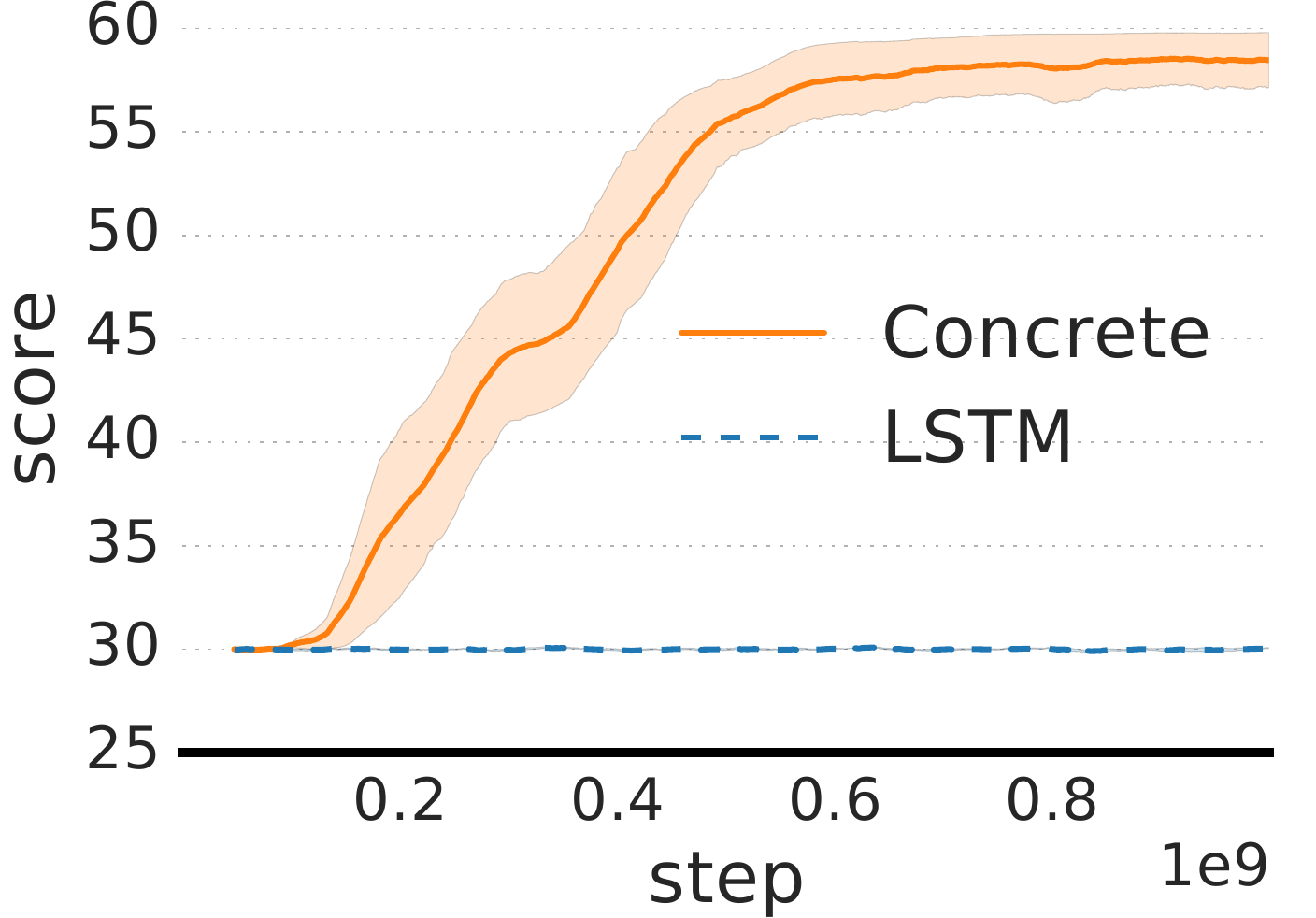}
    \caption*{\em\hspace{4mm}Every 20 steps.}
  \end{subfigure}
  \hspace*{\fill}
  \begin{subfigure}{0.24\linewidth}
    \includegraphics[width=\linewidth]{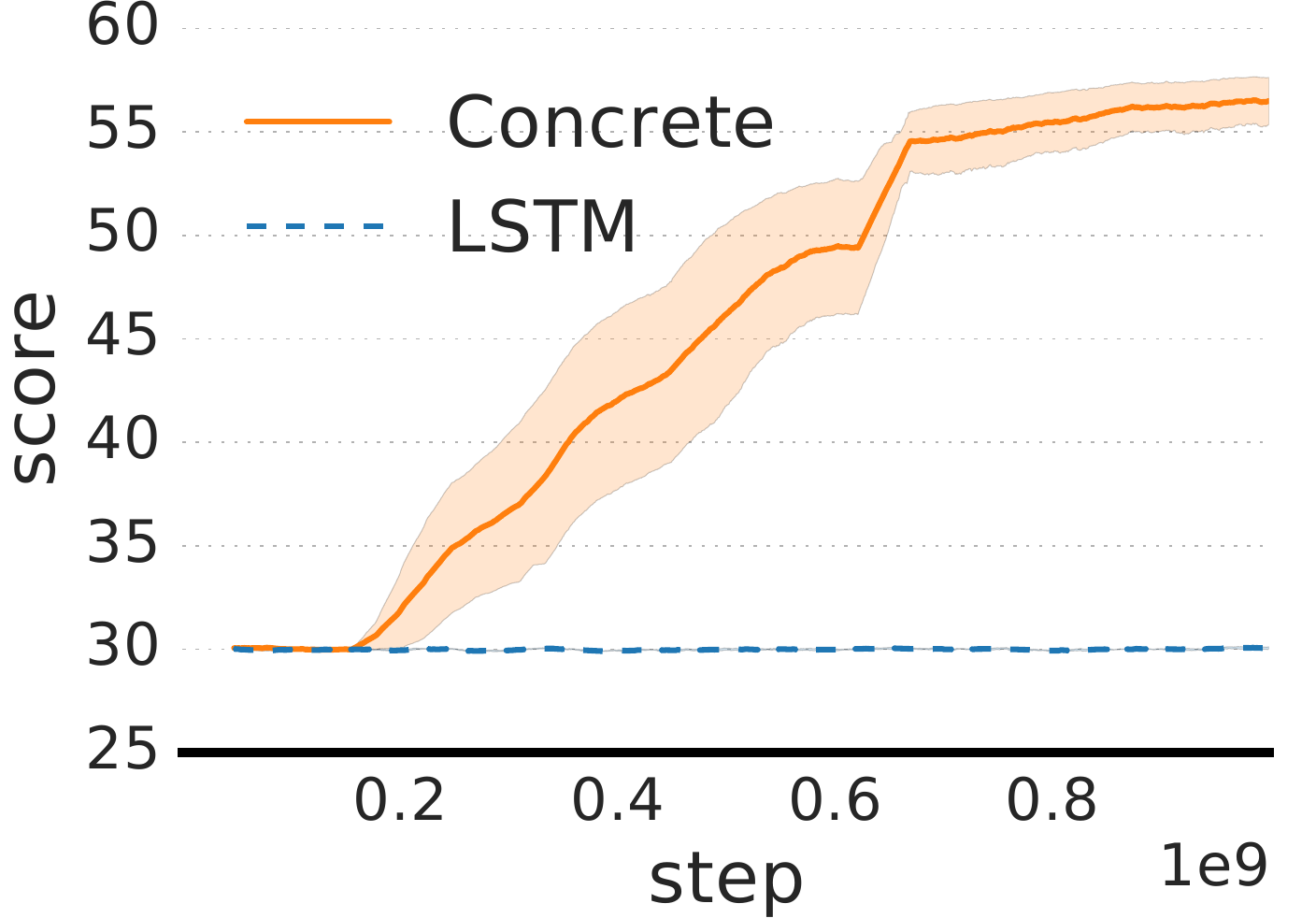}
    \caption*{\em\hspace{4mm}Every 100 steps.}
  \end{subfigure}
  \hspace*{\fill}
  \begin{subfigure}{0.24\linewidth}
    \includegraphics[width=\linewidth]{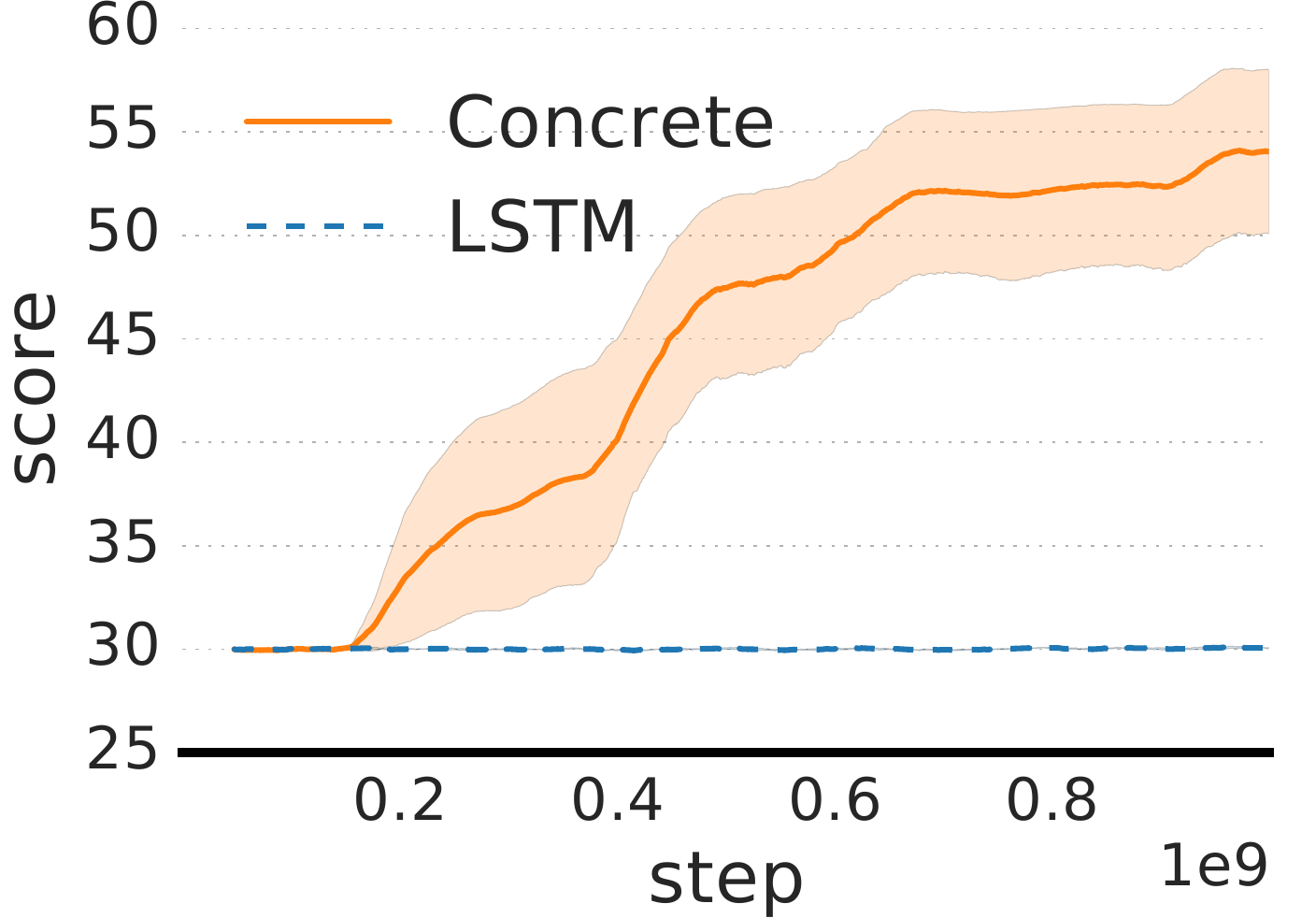}
    \caption*{\em\hspace{4mm}Every 300 steps.}
  \end{subfigure}
  \\[2mm]
  \begin{subfigure}{0.24\linewidth}
    \includegraphics[width=\linewidth]{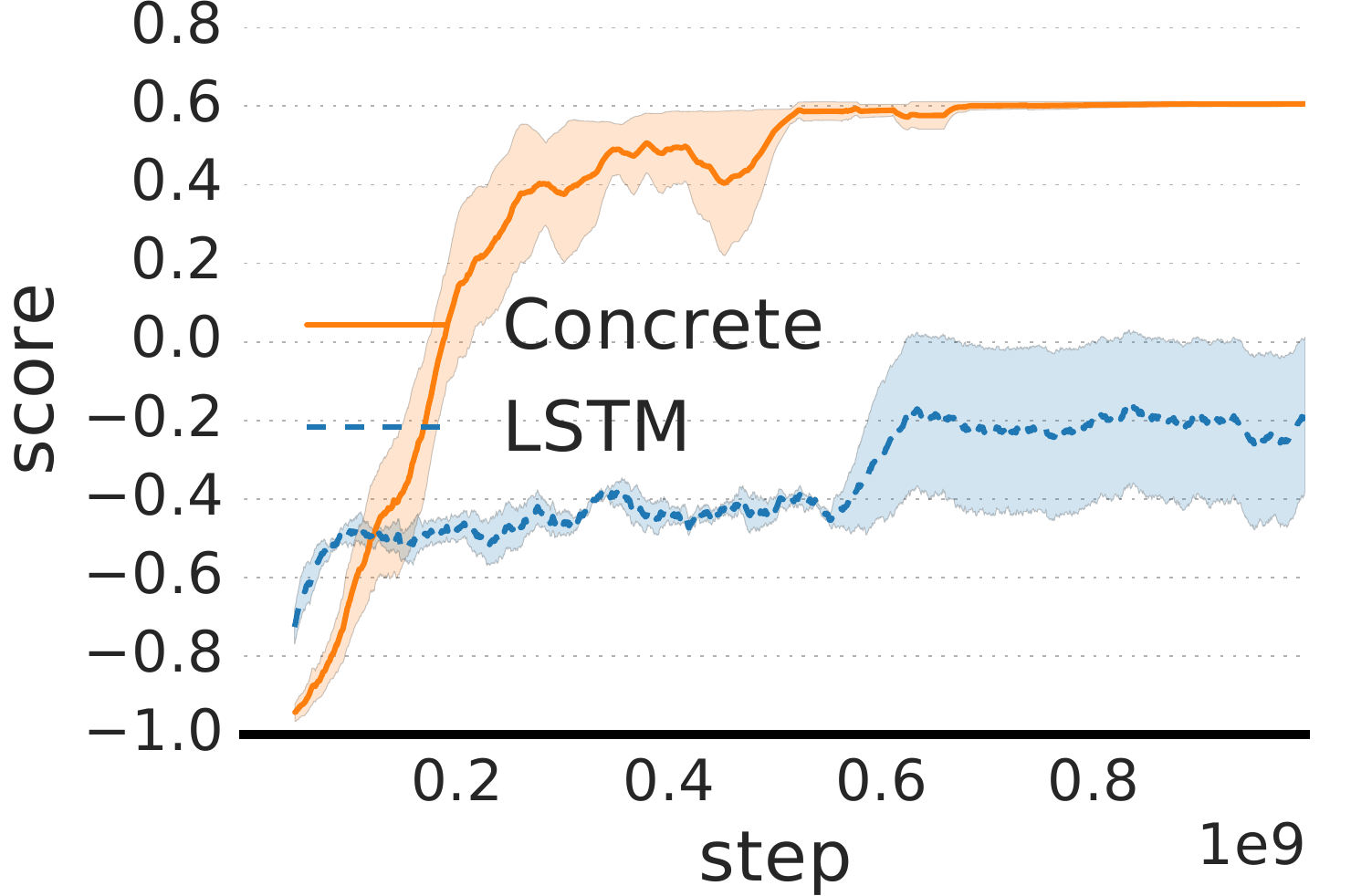}
  \end{subfigure}
  \hspace*{\fill}
  \begin{subfigure}{0.24\linewidth}
    \includegraphics[width=\linewidth]{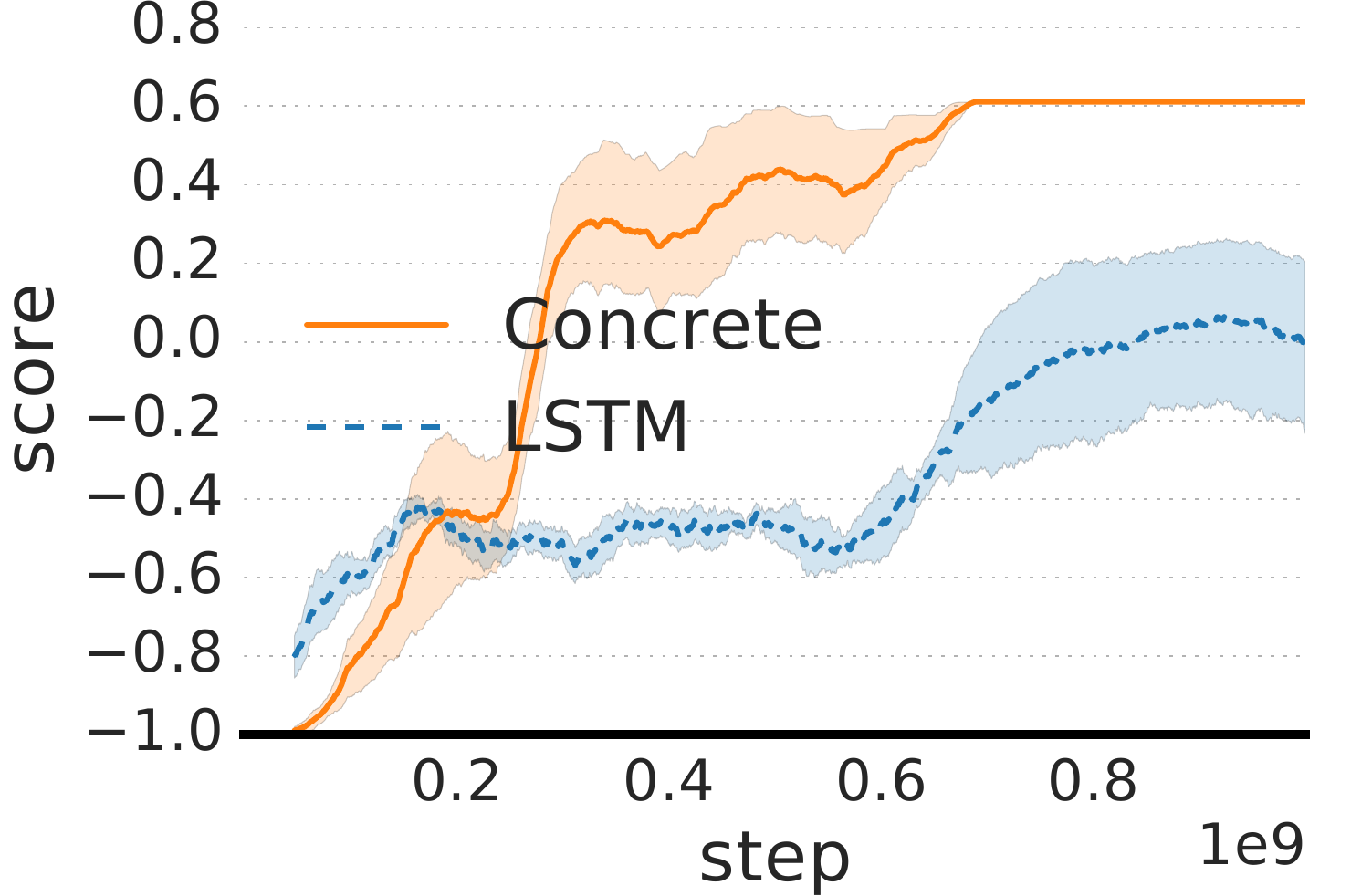}
  \end{subfigure}
  \hspace*{\fill}
  \begin{subfigure}{0.24\linewidth}
    \includegraphics[width=\linewidth]{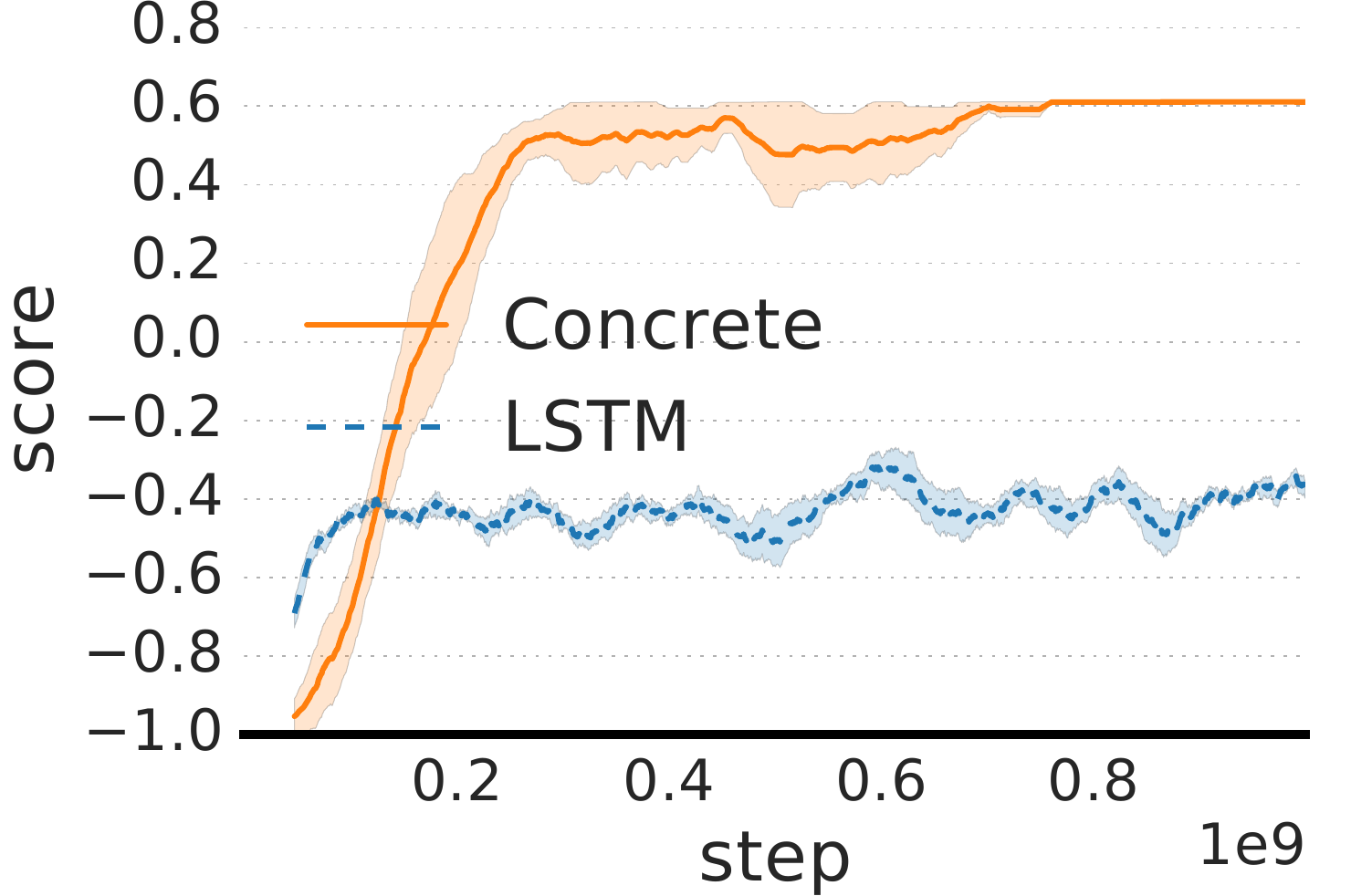}
  \end{subfigure}
  \hspace*{\fill}
  \begin{subfigure}{0.24\linewidth}
    \includegraphics[width=\linewidth]{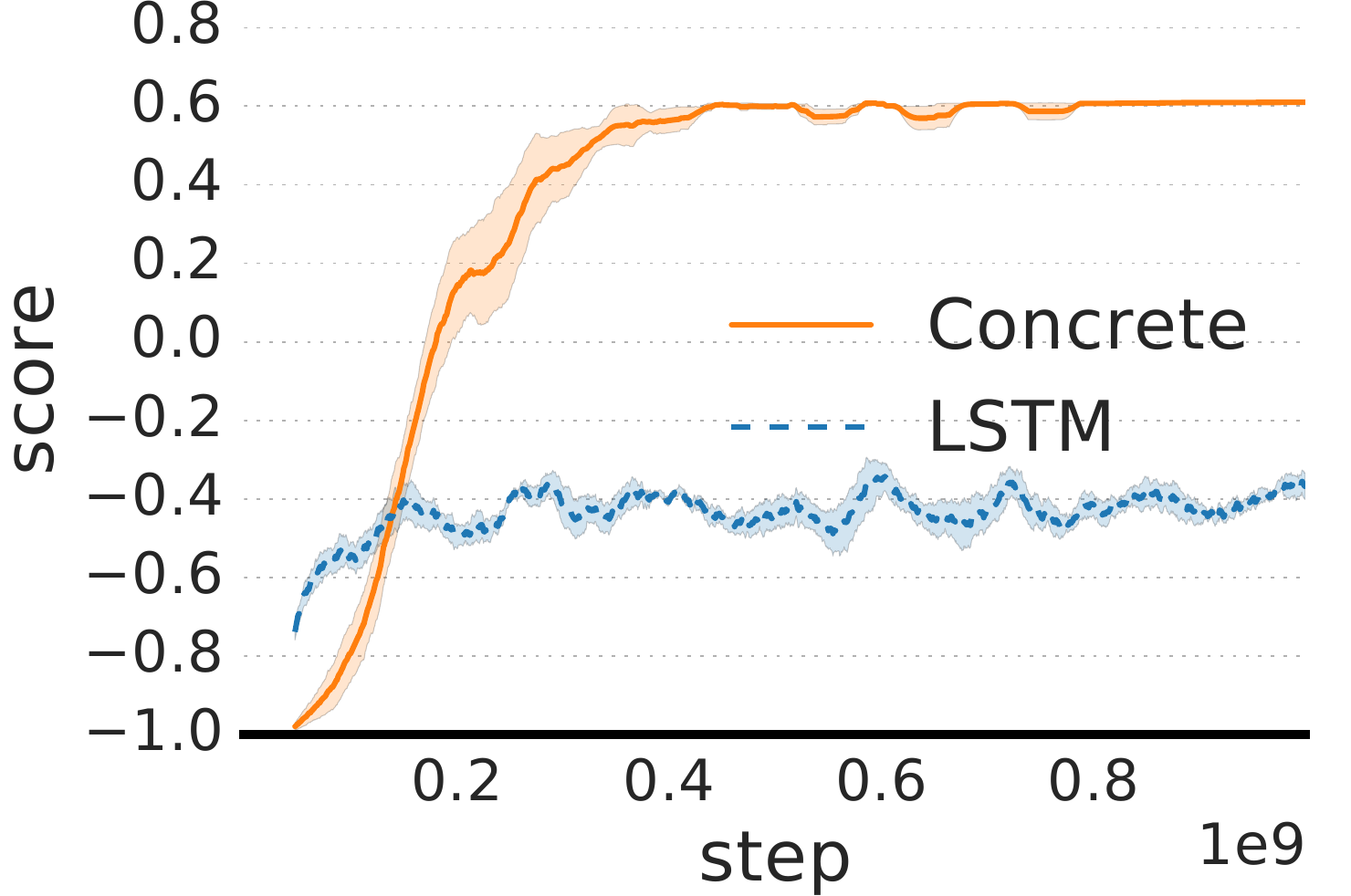}
  \end{subfigure}
  \caption{Performance on the ``Cued Catch'' task with 40 no-reward trials
           followed by 60 trials with reward (top row) and on the ``T-maze''
           task with a 280-timestep detention in ``limbo'' (see Appendix
           sections \ref{app:ccatch}, \ref{app:tmaze}). Orange/solid curves
           depict scores for a network with a Concrete memory, blue/dashed
           curves for a network with an LSTM. The rightmost column shows
           performance for ordinary 300-step network unrolls; preceding
           columns are for the same unrolling but with gradients through the
           hidden state blocked every 100, 20, and 3 steps respectively.
           The best possible score for Cued Catch is 60; for T-maze, just over
           0.6. All curves summarise data from five separate runs of
           population-based training: for each run, the performances of the
           top three seeds in the population are averaged per-step; the
           resulting curve is smoothed, then all five smoothed average
           curves yield the confidence interval plots above. The smoothing
           is cosmetic; for the same plots without smoothing, see
           Appendix secion \ref{app:spcctm}.}
  \label{fig:cctm}
\end{figure*}

``Cued Catch'' is a bandit-like game where the agent must interpret visual
signals to ``catch'' one of two moving blocks. In an episode, the agent
undertakes 100 catching ``trials'', each lasting seven timesteps. The moving
blocks take six steps to reach the agent, and at all times a visual cue
indicating which block to catch shows at the bottom of the screen. There are
four cues, and their associations with either of the two blocks change
with each episode. Episodes start with a pre-trial ``teaching phase'' that
shows correct cue interpretations by pairing cues one-by-one with either of
two additional markers whose associations with the moving blocks are constant
across episodes. Each pairing appears for ten timesteps. For more details and
images of the task, see Appendix section \ref{app:ccatch}.

As described, Cued Catch allows networks using conventional memory architectures
like LSTM to improve gradually in retaining cue semantics: reinforcement
signals from trials just after the teaching phase can ``bootstrap'' the network
into learning to memorise the associations for longer intervals. To disrupt
this bootstrapping, we simply withhold any reward for the first 40 trials. The
agent must retain the associations for at least 280 timesteps before it can
receive any reinforcement for doing so. As shown in Figure \ref{fig:cctm},
this intervention prevents our LSTM-based network from learning the task even
when backpropagating over a generous 300 timesteps; in contrast, the
Concrete-based network learns fairly robustly even with agressively truncated
BPTT.

\subsubsection{T-maze}

In this task, the agent must eventually navigate to one of two possible goal
locations situated at opposite ends of a C-shaped corridor. To learn which goal
to seek, the agent begins the episode in a box-shaped ``teleporter room''
where a relevant visual cue appears. After 50 timesteps, a portion of the room
becomes a teleporter that transfers the agent upon traversal to the C-shaped
corridor after a detention period in ``limbo''---a tiny space where the agent
is unable to move or change its observation in any way. This gives the agent
no option but to store the goal information in memory for a long period.
Appendix section \ref{app:tmaze} presents more details and images of this task.

In our experiments, the detention period in limbo is set to 280 timesteps.
As the agent need remember only one bit of information (compared with the
$\log_2 \binom{4}{2} \approx 2.6$ required for Cued Catch), random
initialisation of large conventional memories may be more likely to start out
with a useful configuration. Nevertheless, as Figure \ref{fig:cctm} shows, only
the Concrete-based network masters this task consistently, again even with
aggressively truncated BPTT.

\subsubsection{Sequence Recall}

\begin{figure}[t]
  \begin{subfigure}{0.49\columnwidth}
    \includegraphics[width=\linewidth]{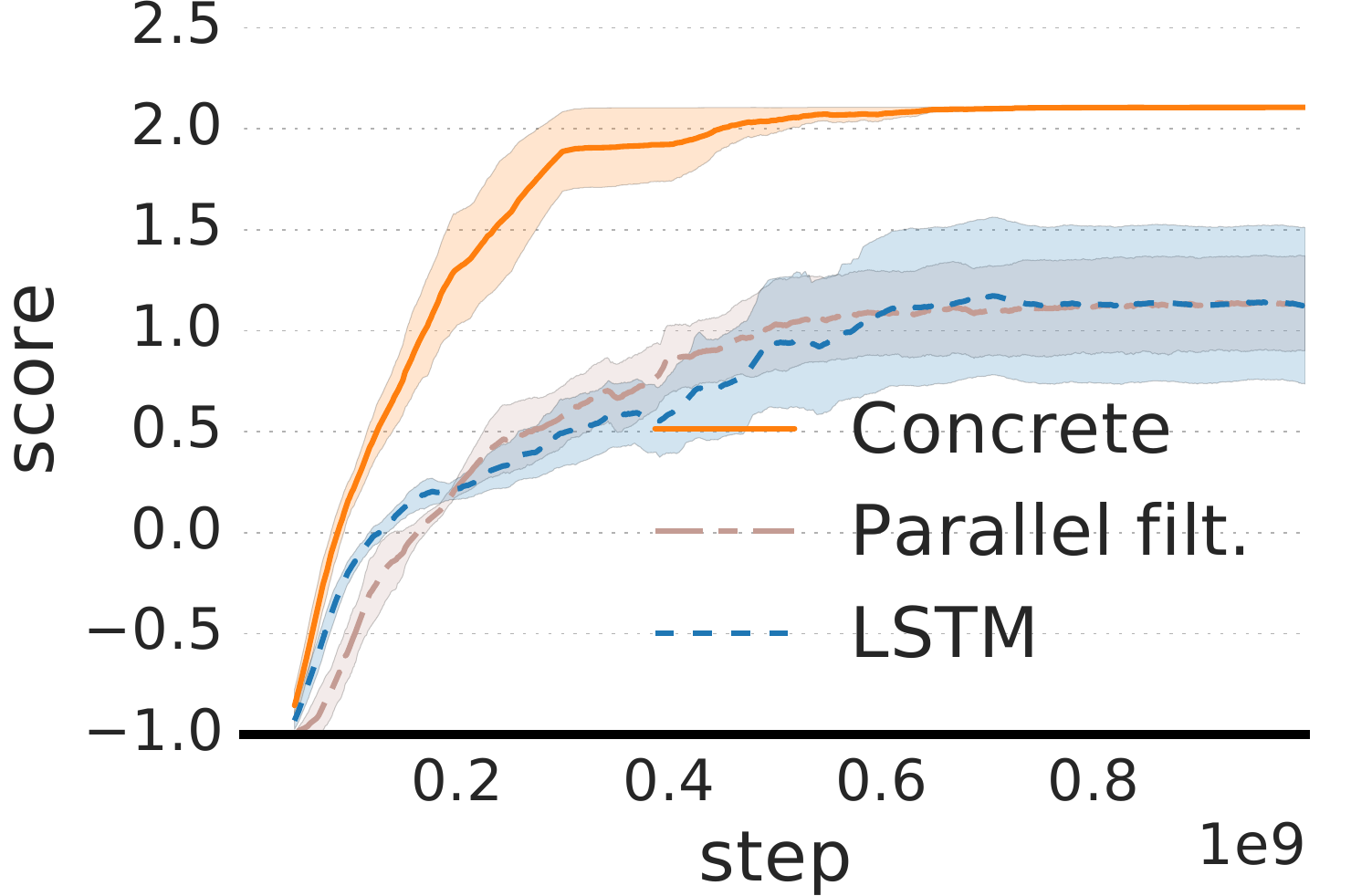}
    \caption*{\em\hspace{4mm}Every 20 steps.}
    \label{fig:srb}
  \end{subfigure}
  \hspace*{\fill}
  \begin{subfigure}{0.49\columnwidth}
    \includegraphics[width=\linewidth]{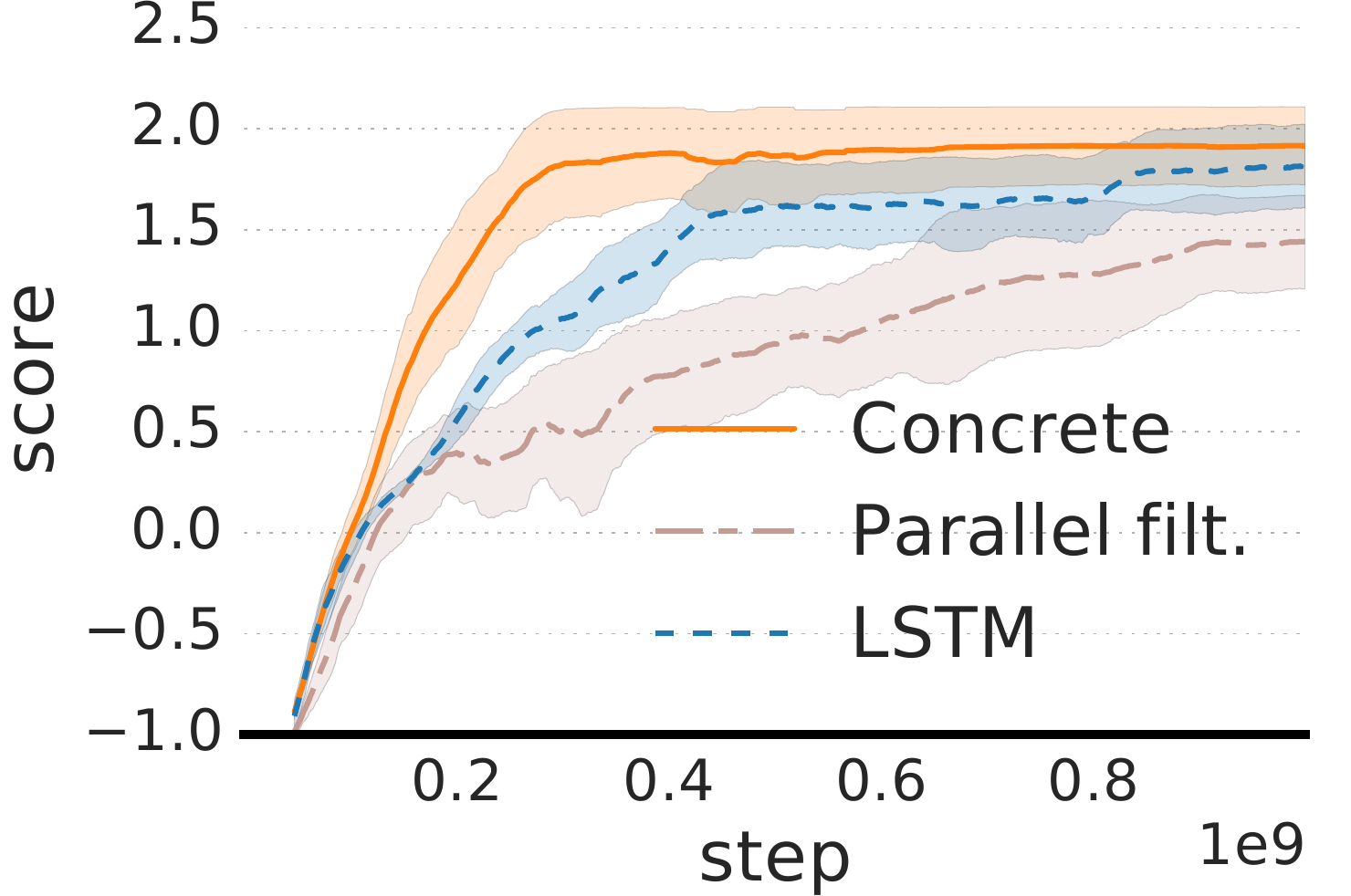}
    \caption*{\em\hspace{4mm}Every 300 steps.}
    \label{fig:sra}
  \end{subfigure}
  \caption{Performance on the ``sequence recall'' task (see Appendix section
           \ref{app:seqrec}); orange/solid curves depict scores for a network
           with a Concrete memory, blue/dashed curves show scores for a
           network with an LSTM, taupe/irregularly-dashed curves show scores
           for a parallel filter-based network like in
           Figure~\ref{fig:actcrit}b.
           Right, for ordinary 300-step network unrolls; left, the same amount
           of unrolling with gradients through the hidden state blocked every
           twenty steps. The best possible score is just over 2.0. The same
           plotting procedure used in Figure \ref{fig:cctm} was applied here;
           for plots without smoothing, see Appendix section \ref{app:spsr}.}
  \label{fig:sr}
\end{figure}

In the ``Sequence Recall'' task, the agent first sits immobilised as it observes
a sequence of four ``light flashes'', each from one of four spatially-separated
disk-shaped ``lights''. Flashes last for 60 timesteps and are separated by
30-timestep gaps. When the sequence ends, the agent is free to walk amongst the
lights. The agent receives a 1.0 reward if the $n$th disk it traverses is the
same as the $n$th light in the sequence. Appendix section \ref{app:seqrec}
presents more details and images of this task.

The sequence recall task requires at least $\log_2 4^4 = 8$ bits of information
storage. Although there are 270 timesteps between the first flash and the
moment when the agent can first move, the agent can apply knowledge of the
last flash relatively soon after it occurs, which may support some
``bootstrapping'' of an effective memory strategy in traditional architectures.
Whether it does nor not, Figure \ref{fig:sr} shows a better performance from the
Concrete-based network over the LSTM baseline. We also applied the ``parallel
filter'' network (Figure~\ref{fig:actcrit}b, Appendix section \ref{app:schem})
to the task; its diminished performance suggests that Concrete's temporal tiling
is useful for tasks where temporal ordering is significant. (We evaluated fewer
BPTT truncation lengths for this task owing to its longer running
time and the extra agent.)

\section{Discussion}

In this work we introduced the LP-RNN, or ``RNN with concrete'', a simple
recurrent memory system that tracks history through a chain of low-pass filter
pools. Unlike popular memory architectures like LSTM, this method
is less sensitive to temporal chunking imposed by truncated backpropagation
through time, a popular gradient method for training recurrent models.

Viewing RNNs with concrete through a signal processing lens revealed the tiled temporal representation benefits of chaining filtering pools over connecting them in parallel.
From a statistical perspective, we discovered parallels to successor representations and found that the linearity of RNNs with concrete---in contrast to the non-linear LSTM---promotes stable representations of the past.

We tested our model on supervised memory tasks and reinforcement learning tasks.
The supervised tasks demonstrate that unlike LSTMs, RNNs with concrete do not require long back-propagation windows to remember distant history.
The reinforcement learning tasks evaluate our model in a high signal-to-noise setting where long term credit assignment is crucial. We showed that RNNs with concrete can retain past information so it can be associated with present context even hundreds of timesteps apart. This is highly valuable in reinforcement learning and often makes the difference between learning well and not learning at all.

\bibliography{references}
\bibliographystyle{icml2018}

\appendix
\onecolumn

\section{Visualisation of data accumulating within a Concrete memory}
\label{app:viz}

The following images depict the evolution over time of the contents of a
six-pool Concrete memory with random orthogonal inter-pool projections. The pool
closest to the input is shown on the left-hand side, and following pools in the
chain extend rightward.

For each image, the first fourteen timesteps (rows) show the contents of all
pools as a one-hot encoding of a 14-character string is fed letter-by-letter
into the memory. Subsequent rows show how the pools change over time with no
further input, or, equivalently, with zero values across all of the input units.
Information can be seen to migrate from a detailed, rapidly changing
representation in the ``upstream'' pools to more slowly changing representations
in the ``downstream'' pools as time goes on. Note how the temporal profile of
these responses match the impulse response curves in Figure \ref{fig:impulse}.

The Concrete implementation used to make these images is an augmented one:
information transferred between pools is also transformed by a random
orthonormal projection and then passed through a $\tanh$ nonlinearity. We allude
to extensions like these in Section \ref{sec:lprnn}. This augmentation has no
particular bearing on the visualisation itself.

Below, the contents of the memory for the string \texttt{machinelearning}:

\includegraphics[width=\linewidth]{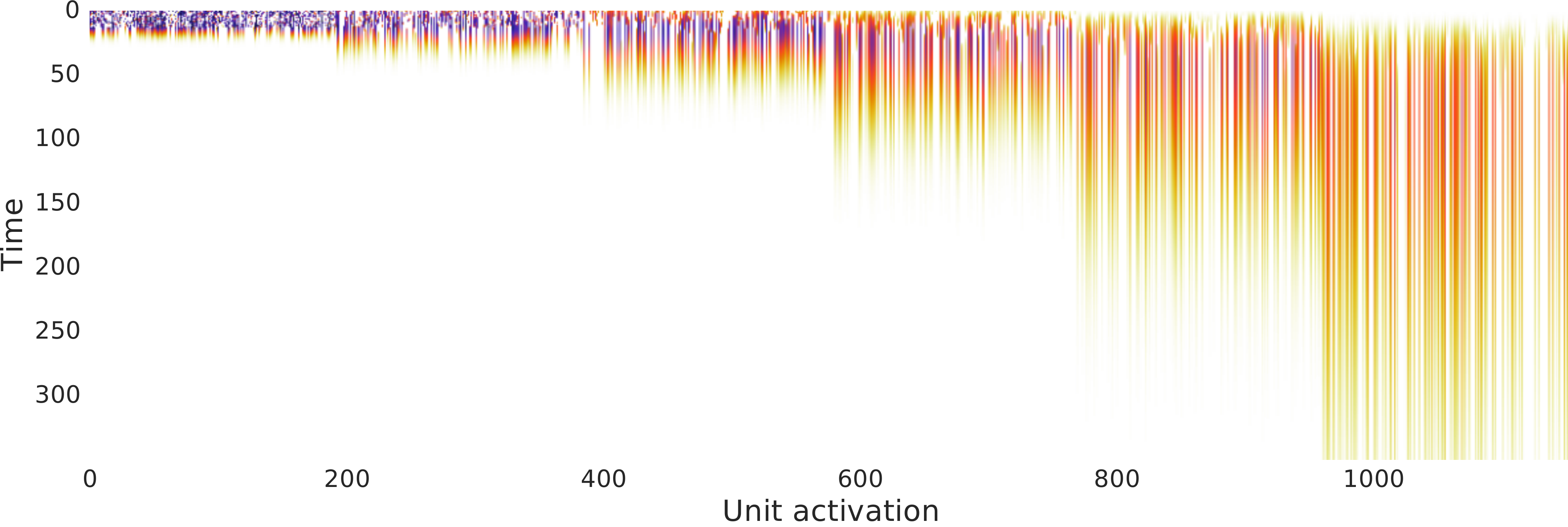}

Next, the same visualisation for the string \texttt{itsmycatmittens}:

\includegraphics[width=\linewidth]{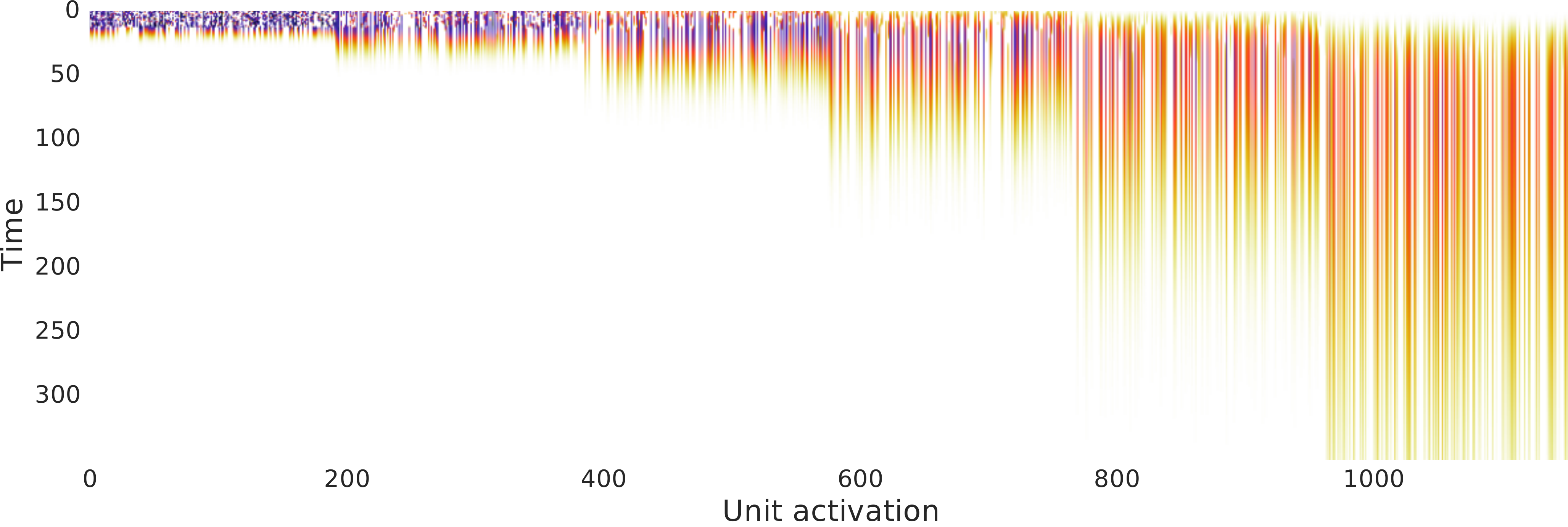}

(Visualisations continue on the following page.)

\newpage

Next, a visualisation of the absolute-valued difference between the two, where
visible differences between both strings are apparent in all pools:

\includegraphics[width=\linewidth]{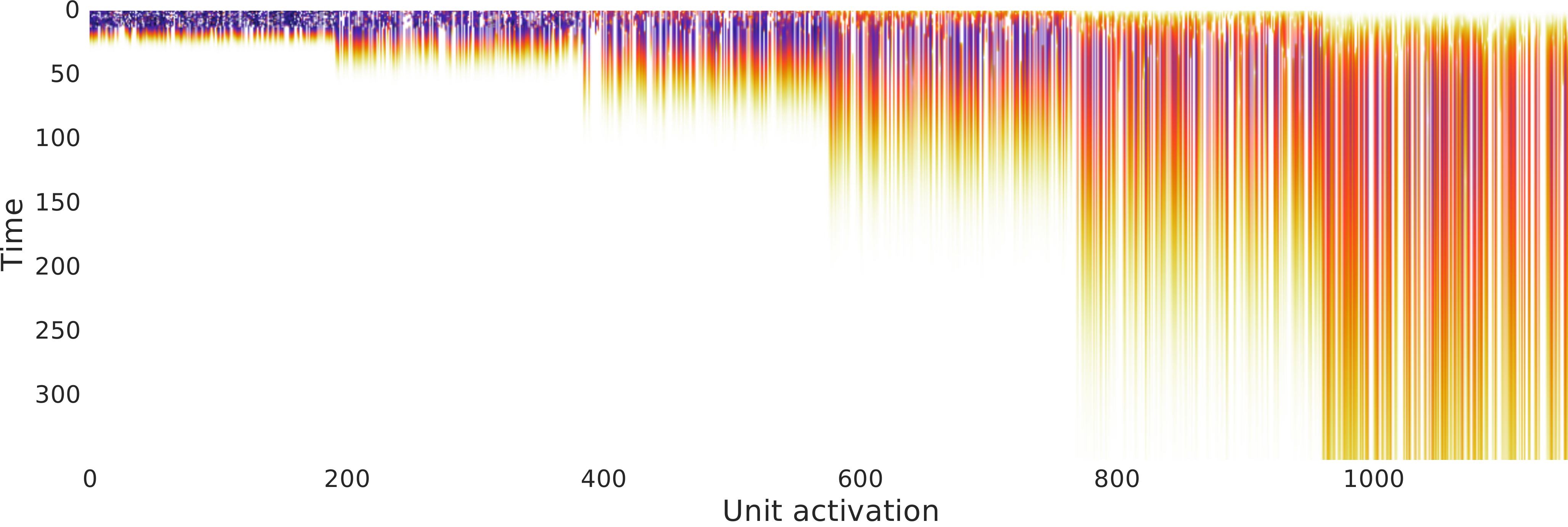}

For distinguishing very similar inputs, it may not suffice to attend to a single
pool. The image below visualises the absolute-valued difference between
the Concrete memory storing the string \texttt{machinelearning} and the same
memory storing an anagram of that string, \texttt{migrainechannel}. White
horizontal bands indicate that there are stretches of time in individual
downstream pools where the representations of the two strings are virtually
indistinguishable. These periods appear not to occur simultaneously across all
pools, however, in which case attention to multiple pools could resolve
ambiguities between the representations.

\includegraphics[width=\linewidth]{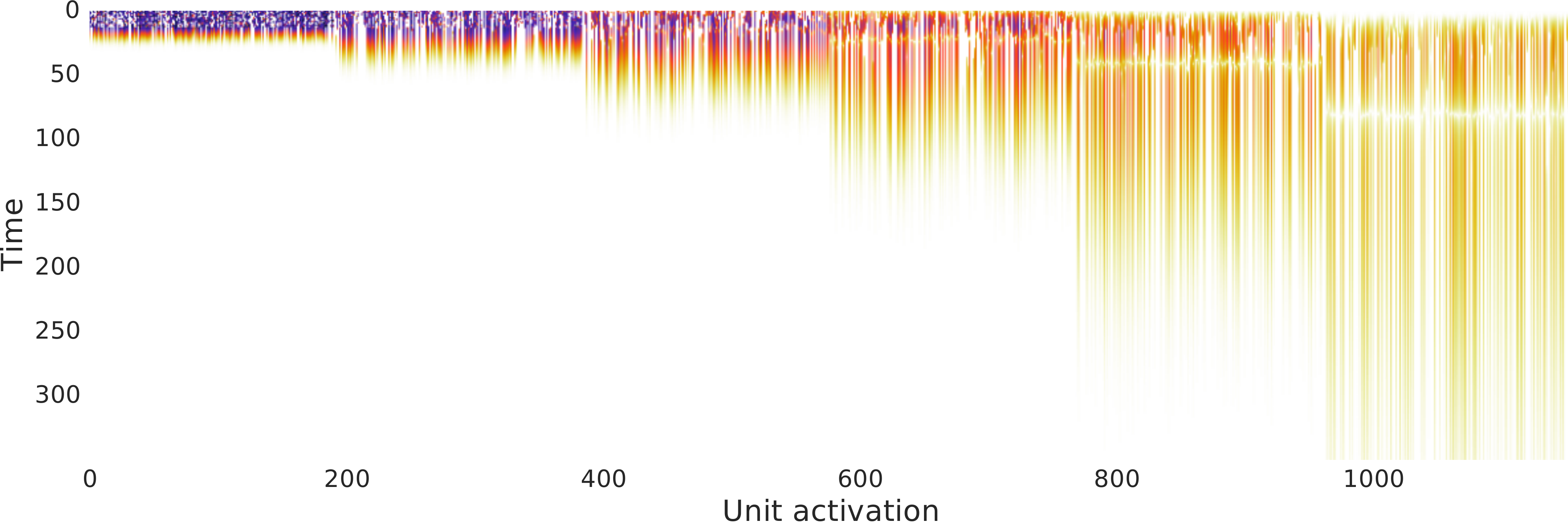}

\section{Descriptions of the RL tasks}

Python implementations of the three RL tasks described below are available at
\url{https://github.com/deepmind/pycolab/tree/master/pycolab/examples}.

\subsection{Cued Catch}
\label{app:ccatch}

\includegraphics[height=0.2\linewidth]{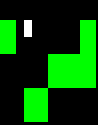}
\hspace{3mm}
\includegraphics[height=0.2\linewidth]{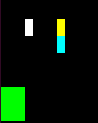}

In this bandit-like game, the mostly-immobile player (single white block) must
``catch'' blocks that repeatedly approach it from the right side of the screen
(yellow and cyan blocks in the right-hand image). The player can only move up
or down between two positions, aligning itself to catch either the yellow or the
the cyan block. The green symbol at the bottom of the screen indicates which
ball to catch; catching this ball earns a reward of 1.0. No other reward is
awarded at any point. A fixed number of these catching trials repeat before the
episode terminates.

There are four ball symbols, two associated with the cyan block and two with the
yellow block. These associations are randomly sampled at the beginning of each
episode. The player is told these associations during a ``teaching phase'' at
the episode's beginning (left-hand image), where the symbols are presented just
beneath either of two wide bars (large green blocks), each of which has a fixed
association with either the cyan or the yellow block. The player must learn
these associations so that it can successfully interpret the training. The
teaching phase is also distinguished by two tall rectangles to the left and
right of the player and by the absence of the yellow and cyan blocks.

The memory demands of the task may be increased by temporarily disabling the 1.0
reward for correct ``catches'' for a specified number of trials after the
``teaching phase''.

\subsection{T-maze}
\label{app:tmaze}

\includegraphics[height=0.2\linewidth]{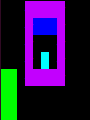}
\hspace{3mm}
\includegraphics[height=0.2\linewidth]{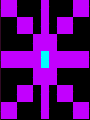}
\hspace{3mm}
\includegraphics[height=0.2\linewidth]{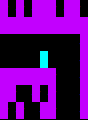}

This version of the classic T-maze memory task places additional demands on the
player's memory by preventing the player from acting on the maze cue for a
configurable number of environment steps.

The episode begins with the player (central cyan block) in a
``teleporter room'' (left image). A cue (tall green rectangle) indicates
whether the player should ultimately seek the goal at the left or right end of
the maze. Eventually, a teleporter appears (blue rectangle above player). Upon
traversal, the player is first teleported to a ``limbo'' (centre image),
where it is completely immobilised. The player is prevented in this way from
using its position in the environment as a means of recording the cue. After a
configurable delay, the player is transported to the horizontal centre of the
T-maze itself, an egocentric scrolling corridor shaped like this:
\rotatebox[origin=c]{90}{\texttt{]}} . The left and right goals are situated
at the lower ends of the two vertical parts of the corridor. Traversing the
goal that matched the initial cue earns a reward of 1.0 for the player;
traversing the wrong goal earns -1.0; either terminates the episode.

An additional penalty of -0.001 is also levied at each timestep (even those
where the player is in ``limbo'') so that waiting for the episode to time out
and terminate is more costly than selecting the wrong goal.

\subsection{Sequence Recall}
\label{app:seqrec}

\includegraphics[height=0.2\linewidth]{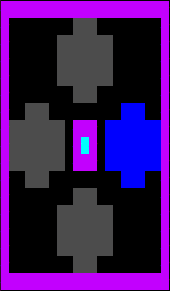}
\hspace{3mm}
\includegraphics[height=0.2\linewidth]{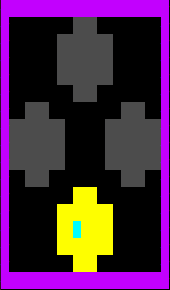}

This task partly resembles the operation of an electronic memory game toy.
At the beginning of the episode (left image), the player (centre cyan block)
sits immobilised within a small ``pen'' surrounded by four disk-shaped
``lights''. A random sequence of four lights (each drawn with replacement) is
shown (in the left-hand image, the blue, rightmost light is ``on''). After the
light sequence is presented, the player is free to move about the environment
(right-hand image). The player receives a 1.0 reward if the $n$th light it
traverses is the same as the $n$th light in the original sequence, so it must
traverse the lights in the same order in which they were presented in order to
achieve the highest score. If the same light appears in two adjacent positions
in the sequence, the player must enter, leave, and re-enter the light disk in
order to receive credit for both positions.

Similar to the T-maze task, an additional penalty of -0.005 is levied at each
timestep so that waiting for the episode to time out and terminate is more
costly than traversing four incorrect lights.

\section{Supplemental plots}

\subsection{Sequence classification performance by minibatch size}
\label{app:suppsc}

\begin{figure}[H]
  \begin{subfigure}{0.325\linewidth}
    \includegraphics[width=\linewidth]{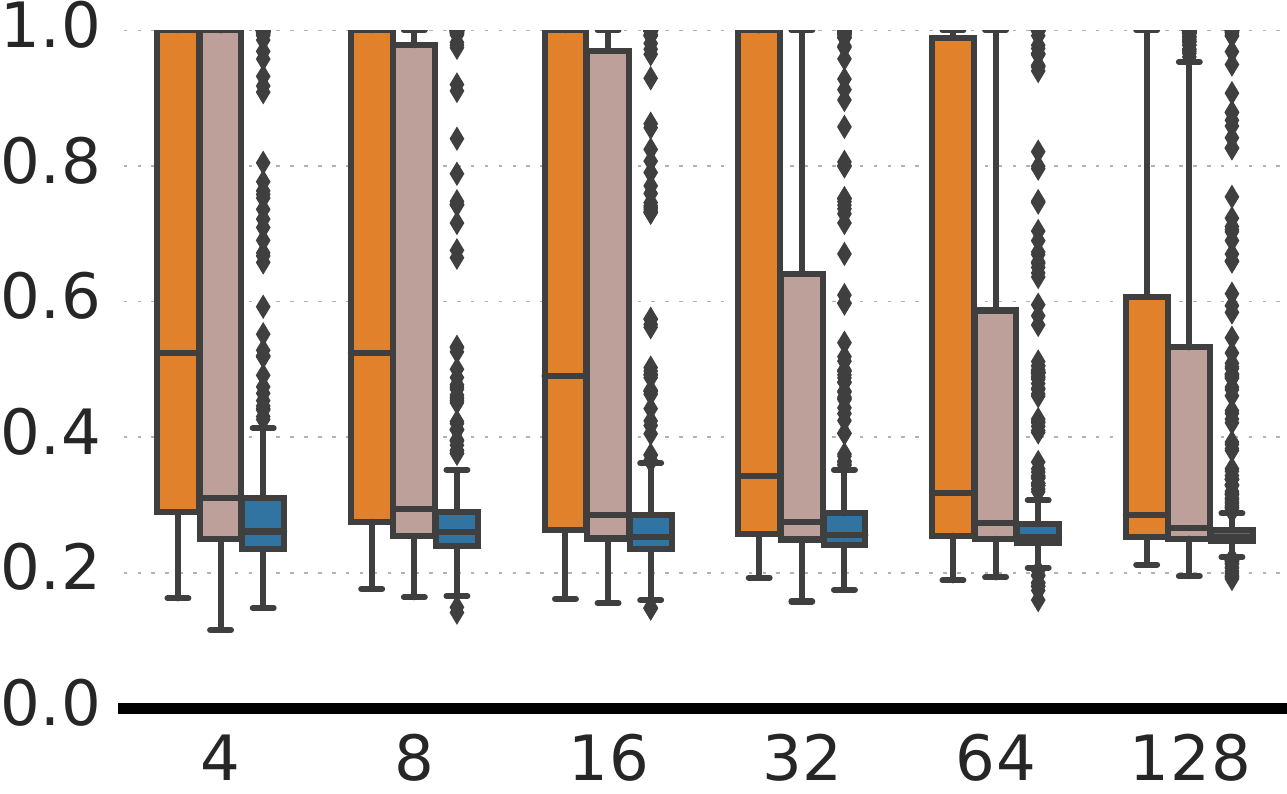}
    \caption{Two markers.}
    \label{fig:suppsca}
  \end{subfigure}
  \hspace*{\fill}
  \begin{subfigure}{0.325\linewidth}
    \includegraphics[width=\linewidth]{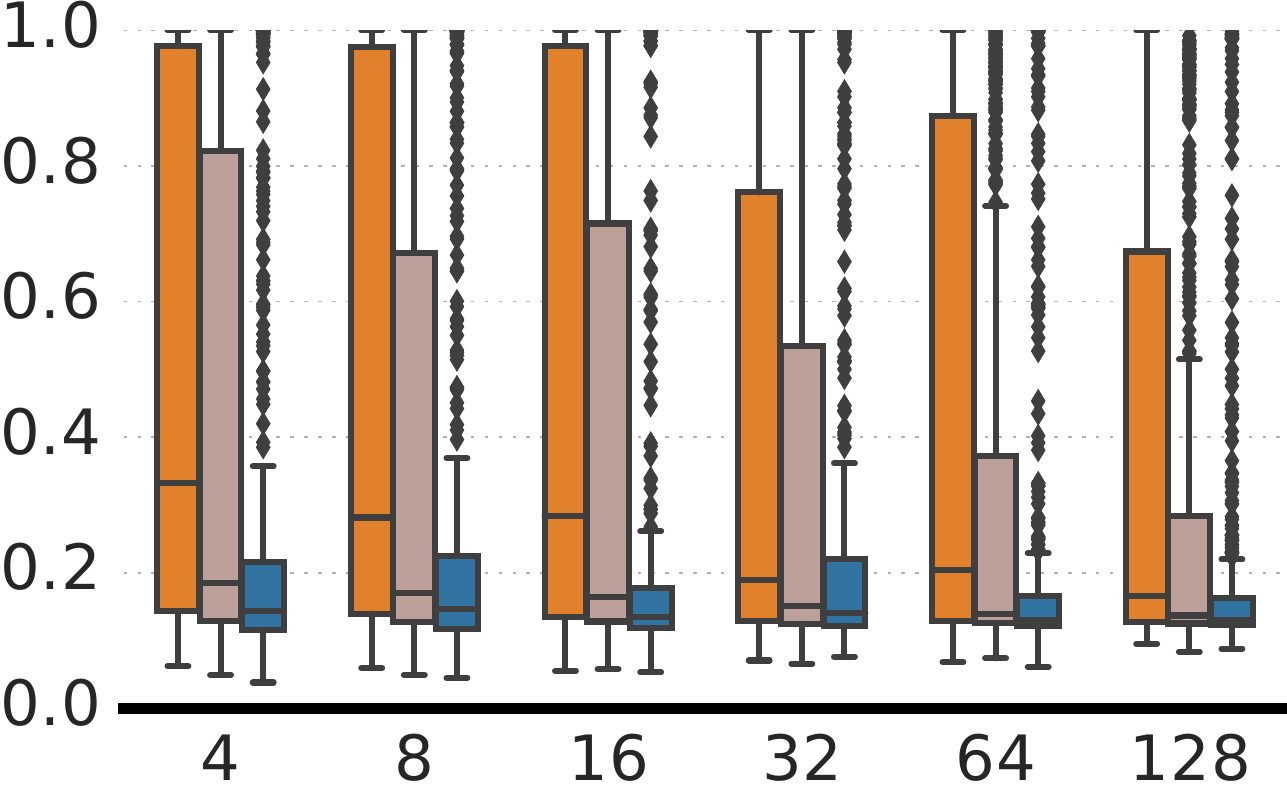}
    \caption{Three markers.}
    \label{fig:suppscb}
  \end{subfigure}
  \hspace*{\fill}
  \begin{subfigure}{0.325\linewidth}
    \includegraphics[width=\linewidth]{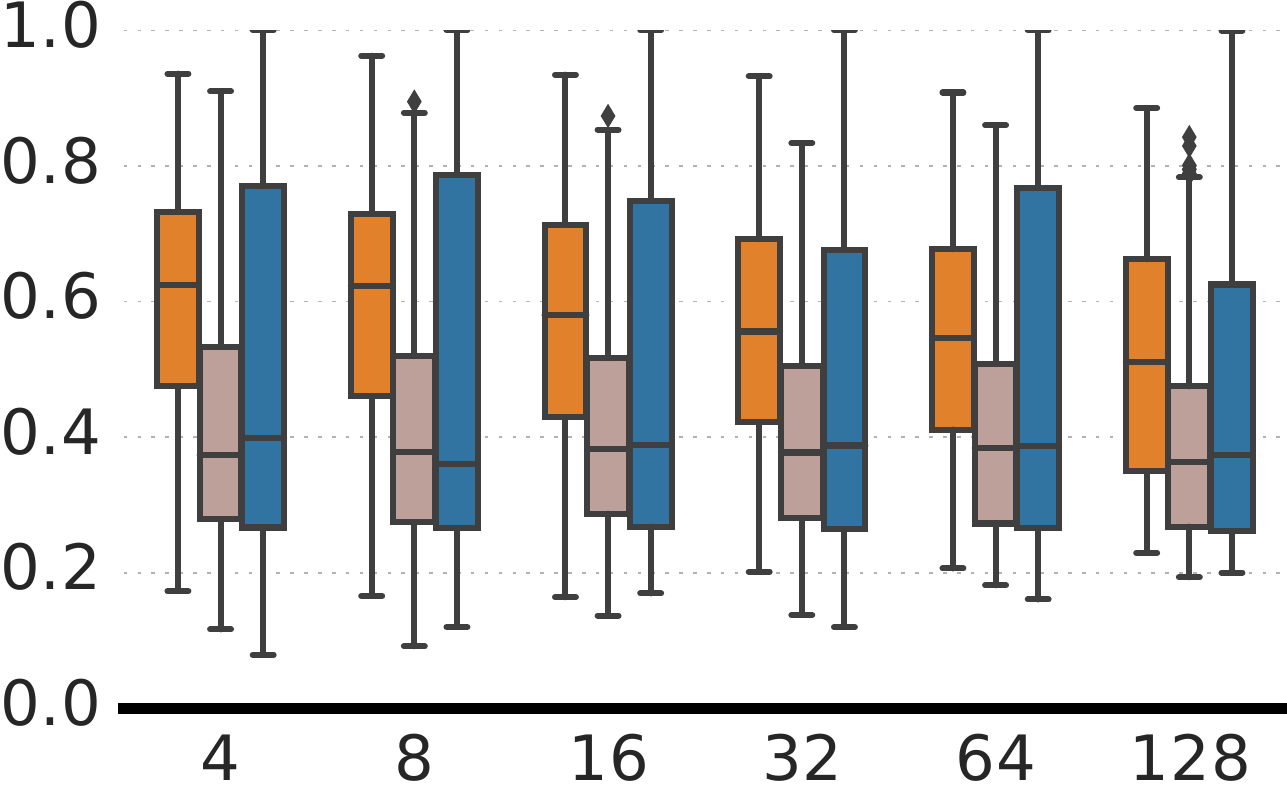}
    \caption{Two subsequence markers.}
    \label{fig:suppscc}
  \end{subfigure} \\[2mm]
  \includegraphics[width=0.45\linewidth]{temporal_order_legend.pdf}
  \caption{As a supplement to the discussion surrounding Figure \ref{fig:sc},
           sequence classification performance sliced by minibatch size.
           The low-pass RNN-based methods achieve better results with smaller
           minibatches. This effect, along with the poorer performance for
           longer truncation intervals observed in Figure \ref{fig:sc}, may be
           be an instance of the infrequent update problem discussed in the
           introduction: for the largest batch size (128) and truncation
           interval (256), the entire training for the original sequence
           classification tasks will apply only 1,220 weight updates to the
           network. For the smallest batch size (4) and truncation interval
           (2), there will be five million updates.}
\end{figure}

\subsection{Plots from Figure \ref{fig:cctm} without smoothing}
\label{app:spcctm}

\begin{figure}[H]
  \begin{subfigure}{0.24\linewidth}
    \includegraphics[width=\linewidth]{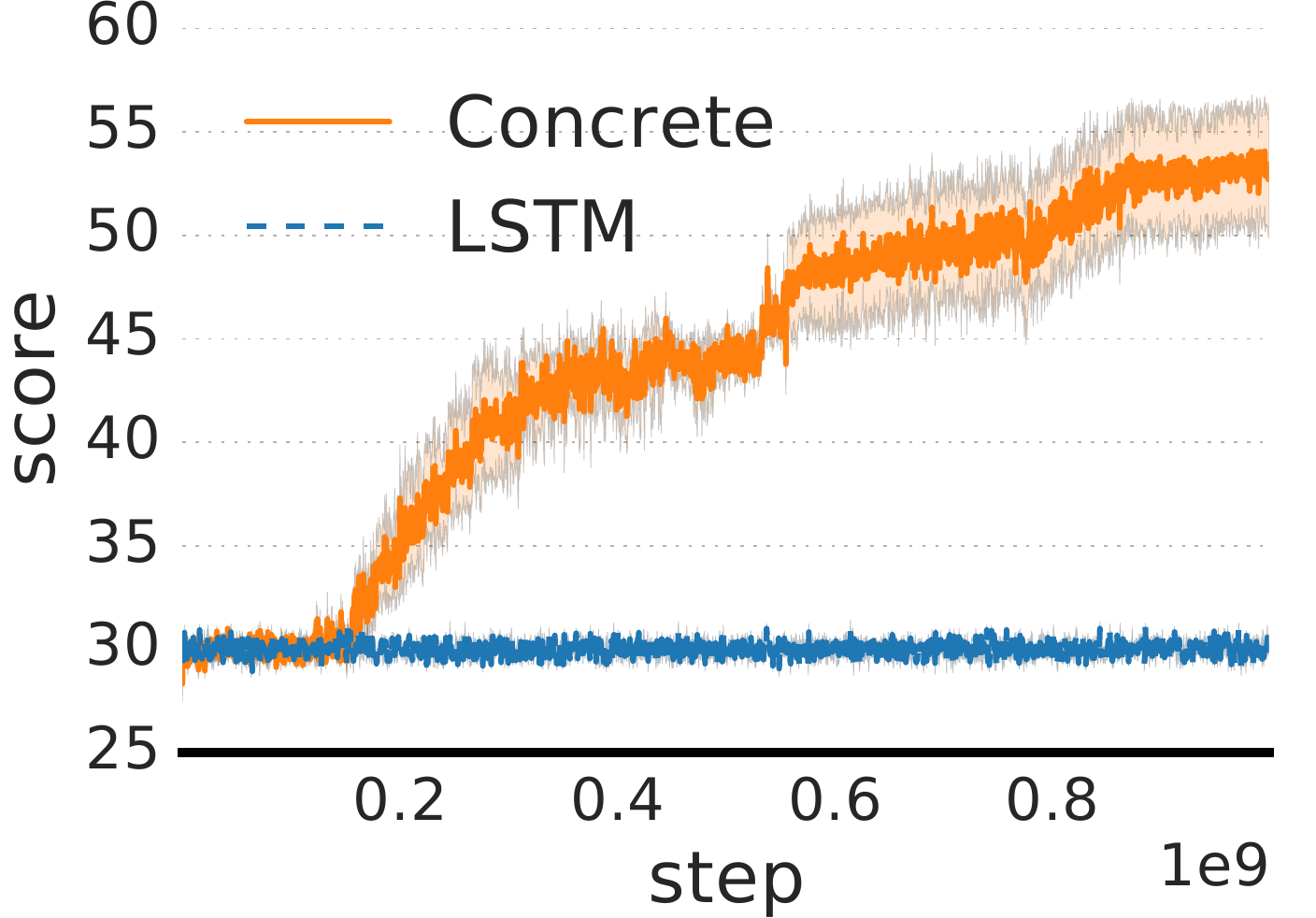}
    \caption*{\em\hspace{4mm}Every 3 steps.}
  \end{subfigure}
  \hspace*{\fill}
  \begin{subfigure}{0.24\linewidth}
    \includegraphics[width=\linewidth]{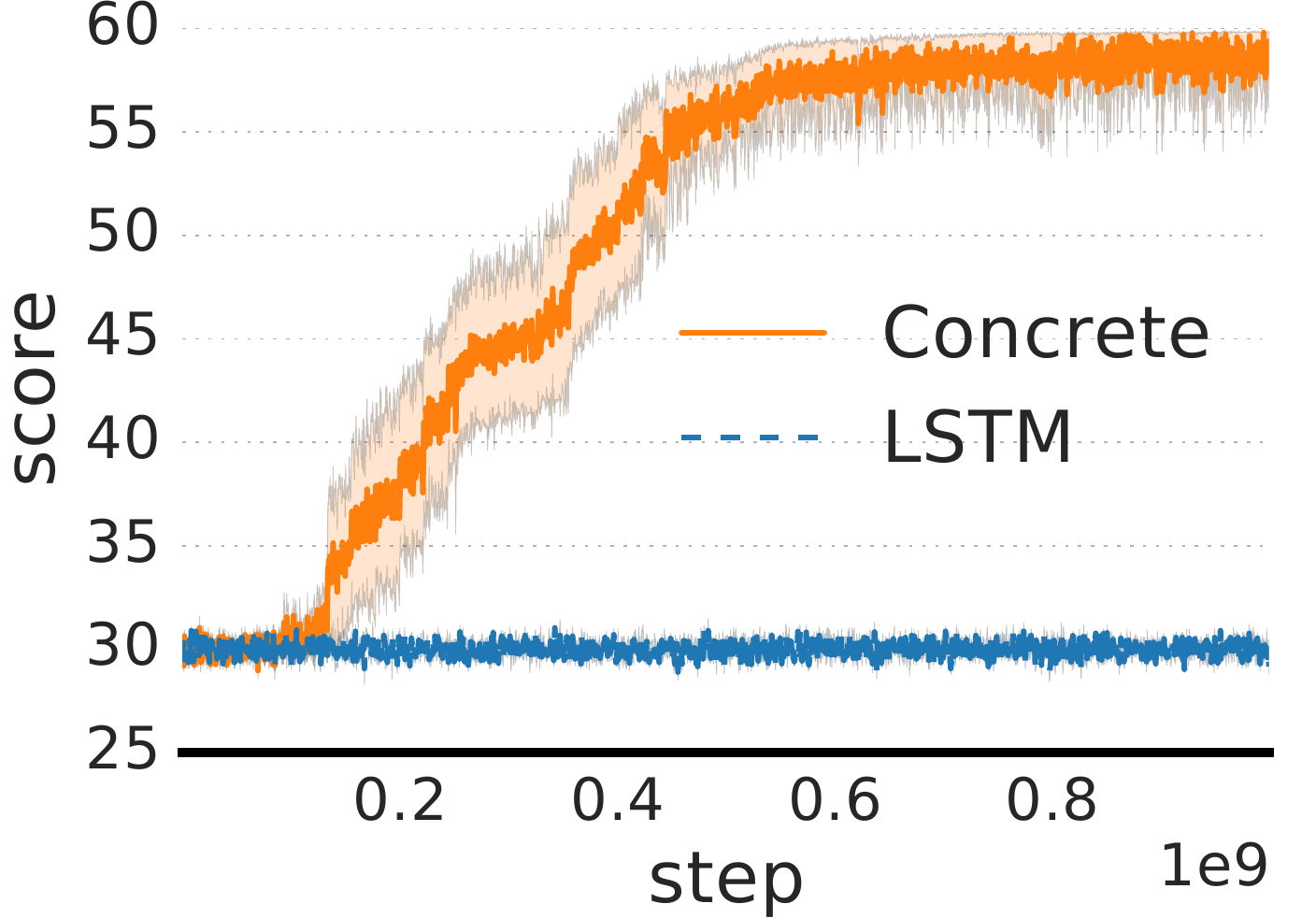}
    \caption*{\em\hspace{4mm}Every 20 steps.}
  \end{subfigure}
  \hspace*{\fill}
  \begin{subfigure}{0.24\linewidth}
    \includegraphics[width=\linewidth]{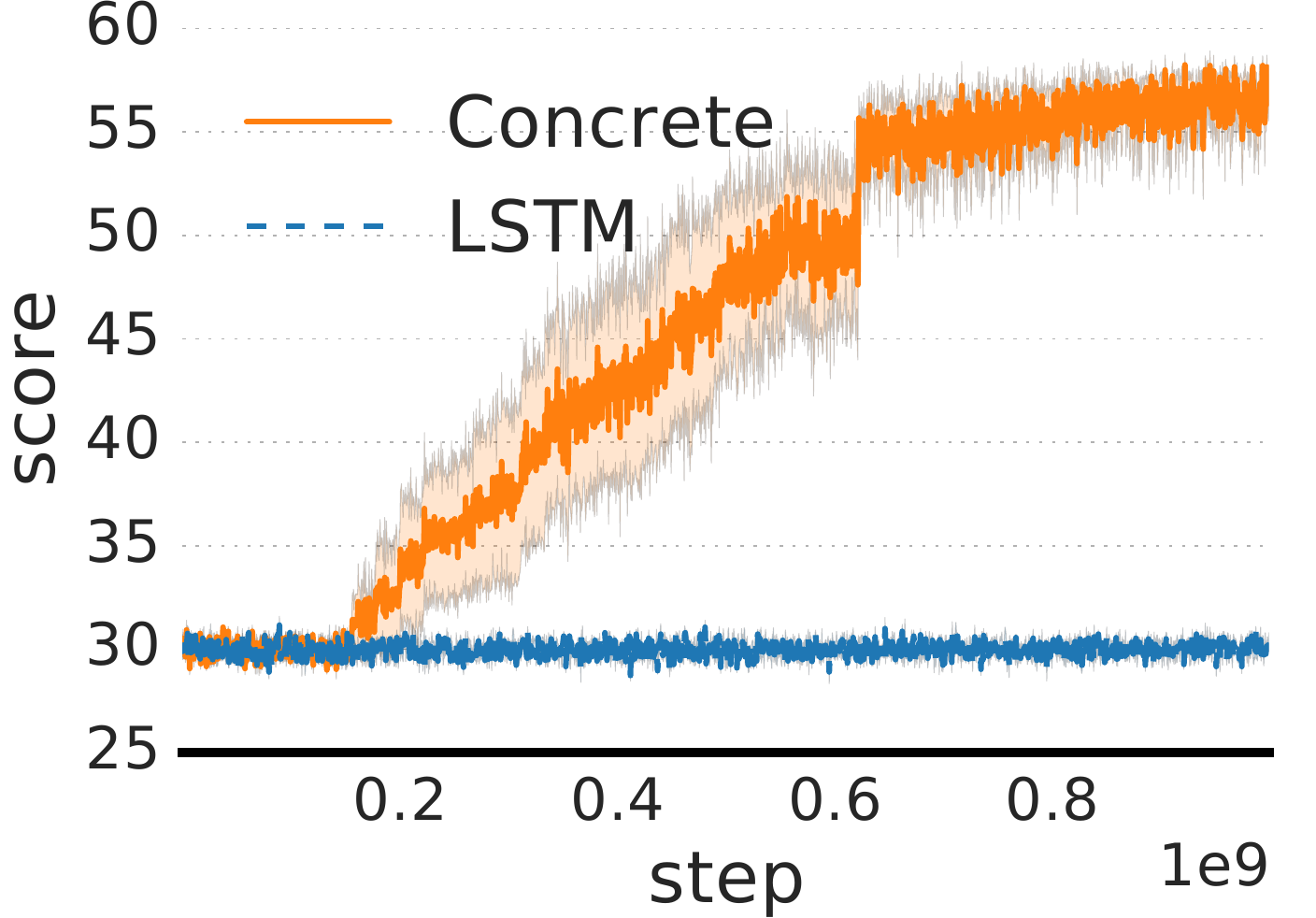}
    \caption*{\em\hspace{4mm}Every 100 steps.}
  \end{subfigure}
  \hspace*{\fill}
  \begin{subfigure}{0.24\linewidth}
    \includegraphics[width=\linewidth]{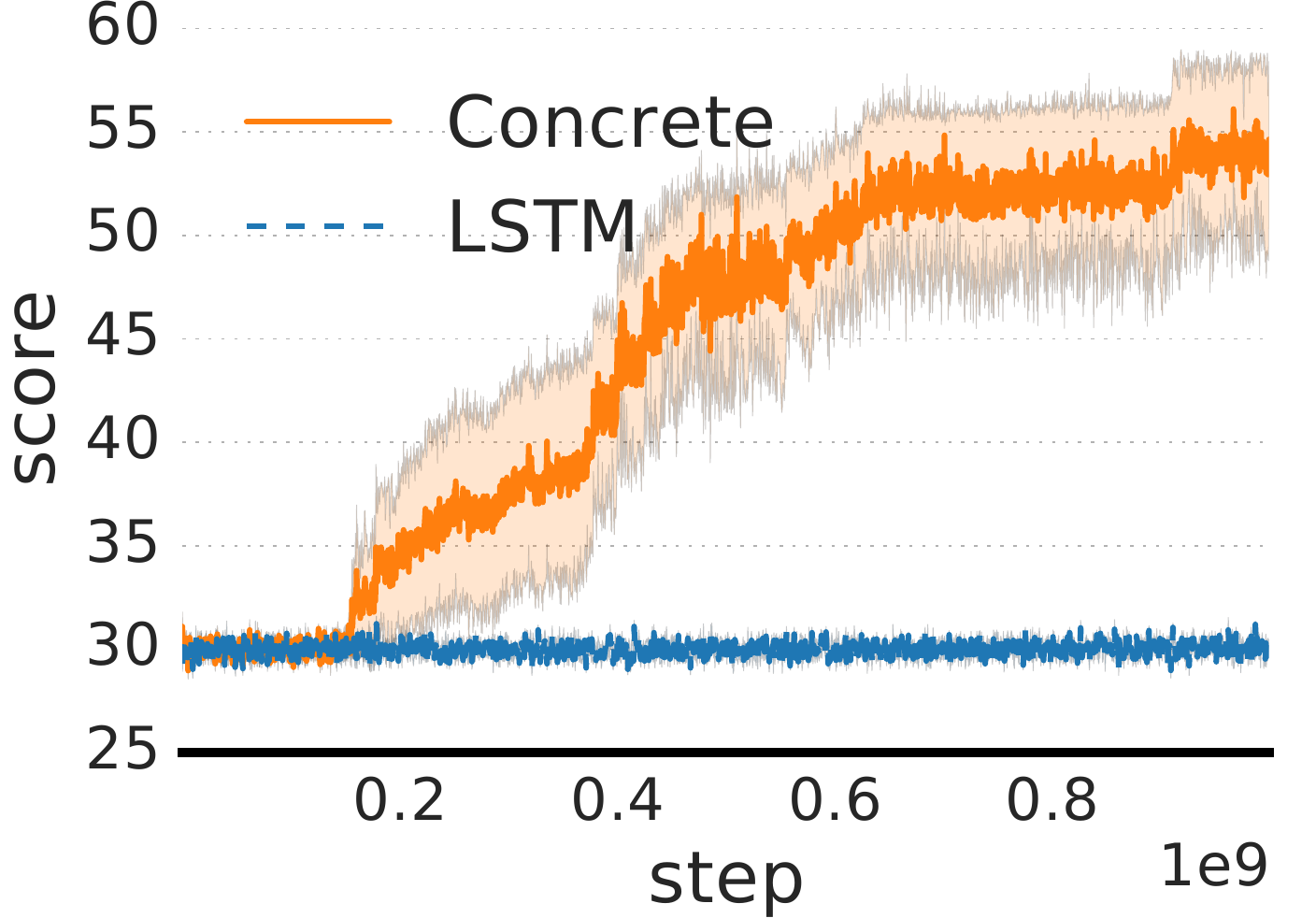}
    \caption*{\em\hspace{4mm}Every 300 steps.}
  \end{subfigure}
  \\[2mm]
  \begin{subfigure}{0.24\linewidth}
    \includegraphics[width=\linewidth]{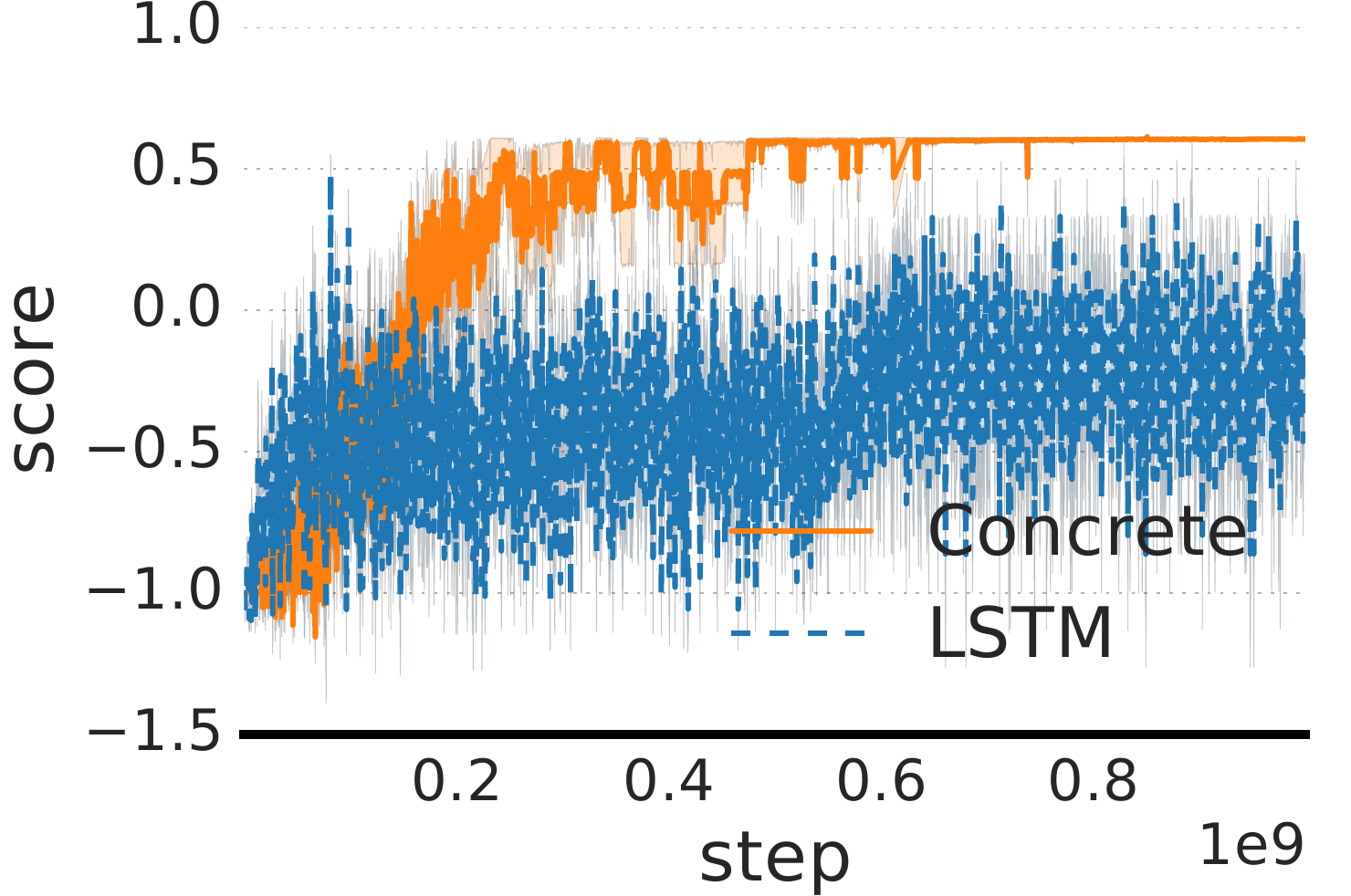}
  \end{subfigure}
  \hspace*{\fill}
  \begin{subfigure}{0.24\linewidth}
    \includegraphics[width=\linewidth]{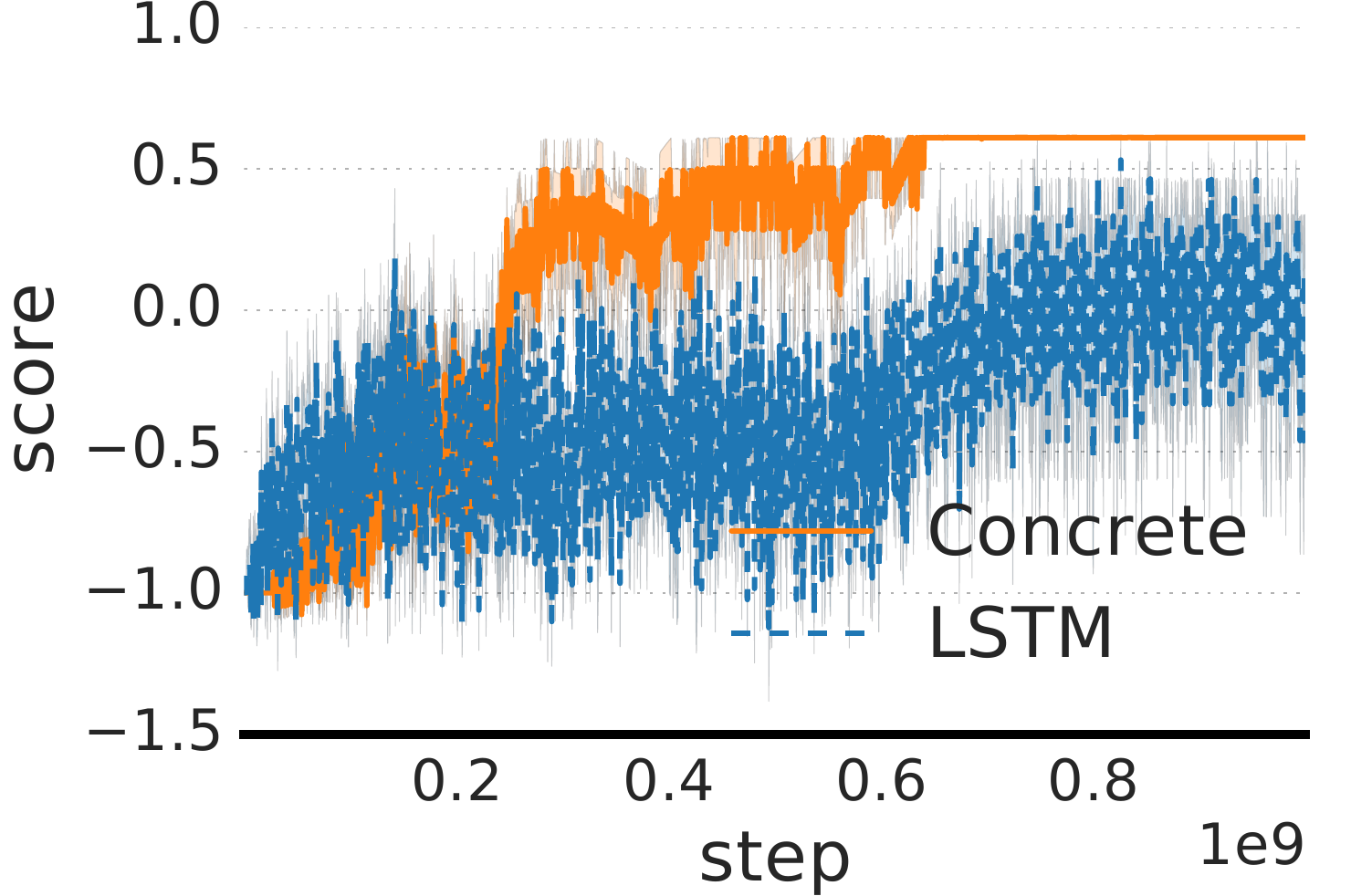}
  \end{subfigure}
  \hspace*{\fill}
  \begin{subfigure}{0.24\linewidth}
    \includegraphics[width=\linewidth]{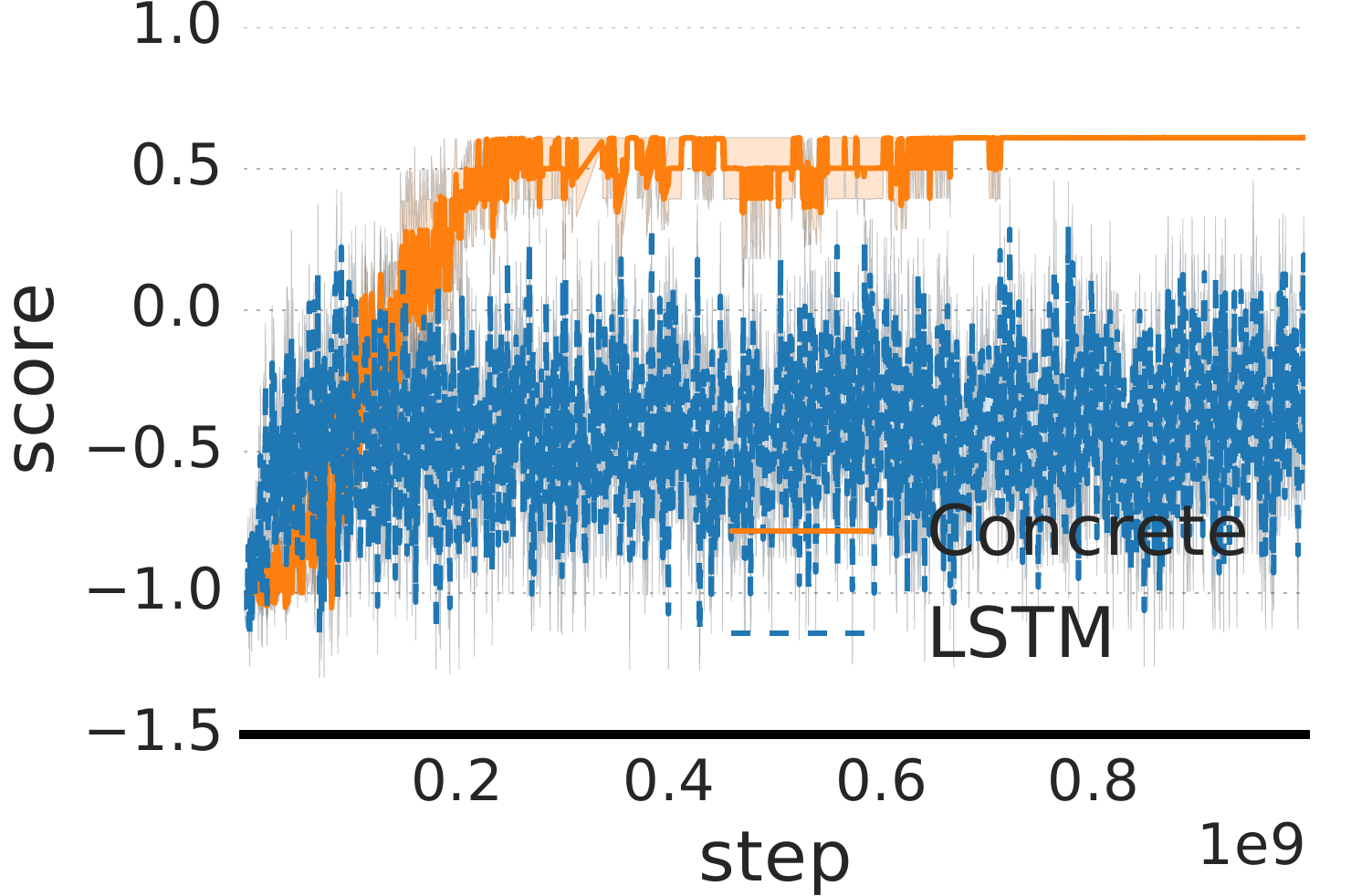}
  \end{subfigure}
  \hspace*{\fill}
  \begin{subfigure}{0.24\linewidth}
    \includegraphics[width=\linewidth]{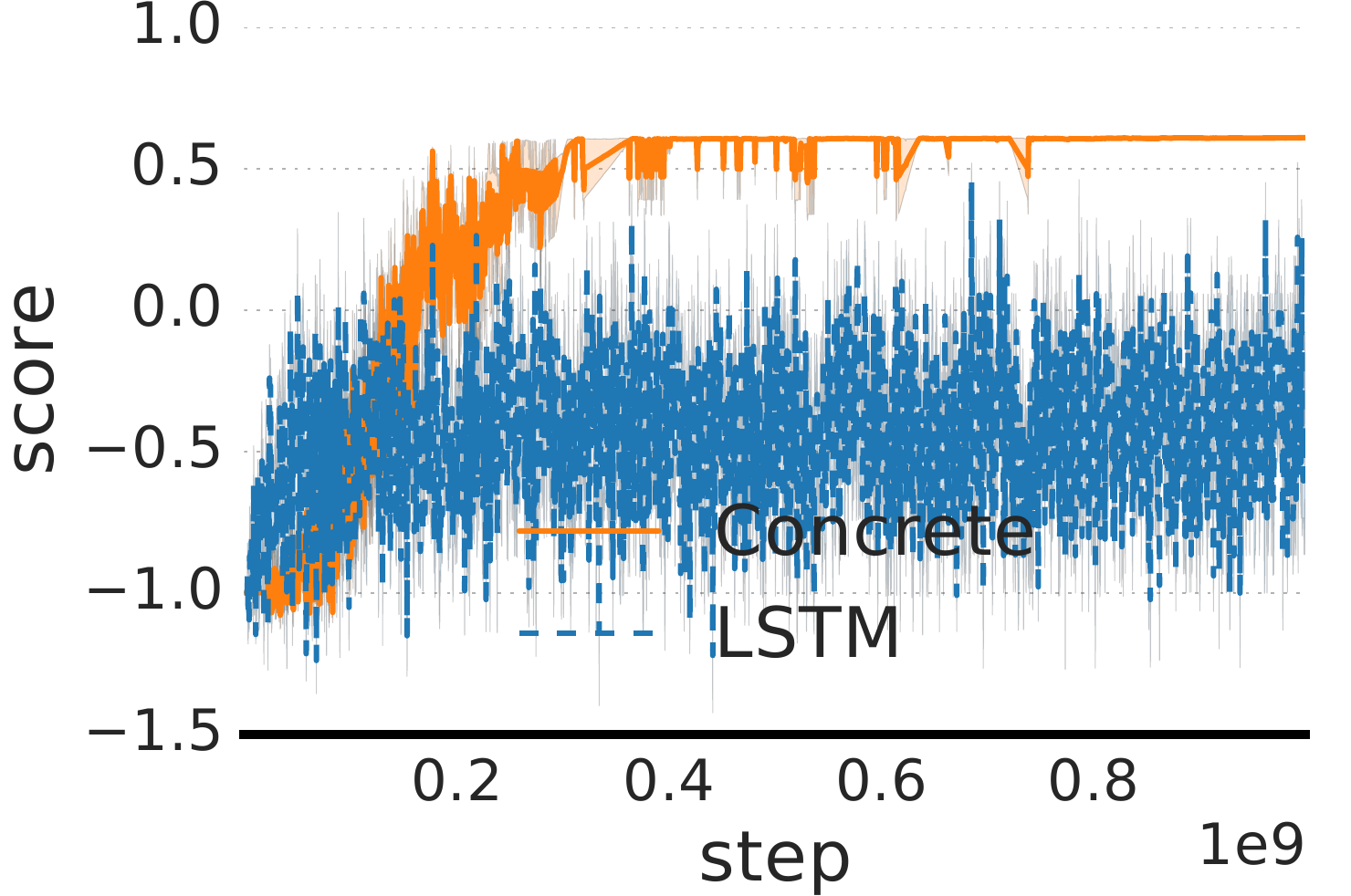}
  \end{subfigure}
\end{figure}

\subsection{Plots from Figure \ref{fig:sr} without smoothing}
\label{app:spsr}

\begin{figure}[H]
  \begin{subfigure}{0.24\linewidth}
    \includegraphics[width=\linewidth]{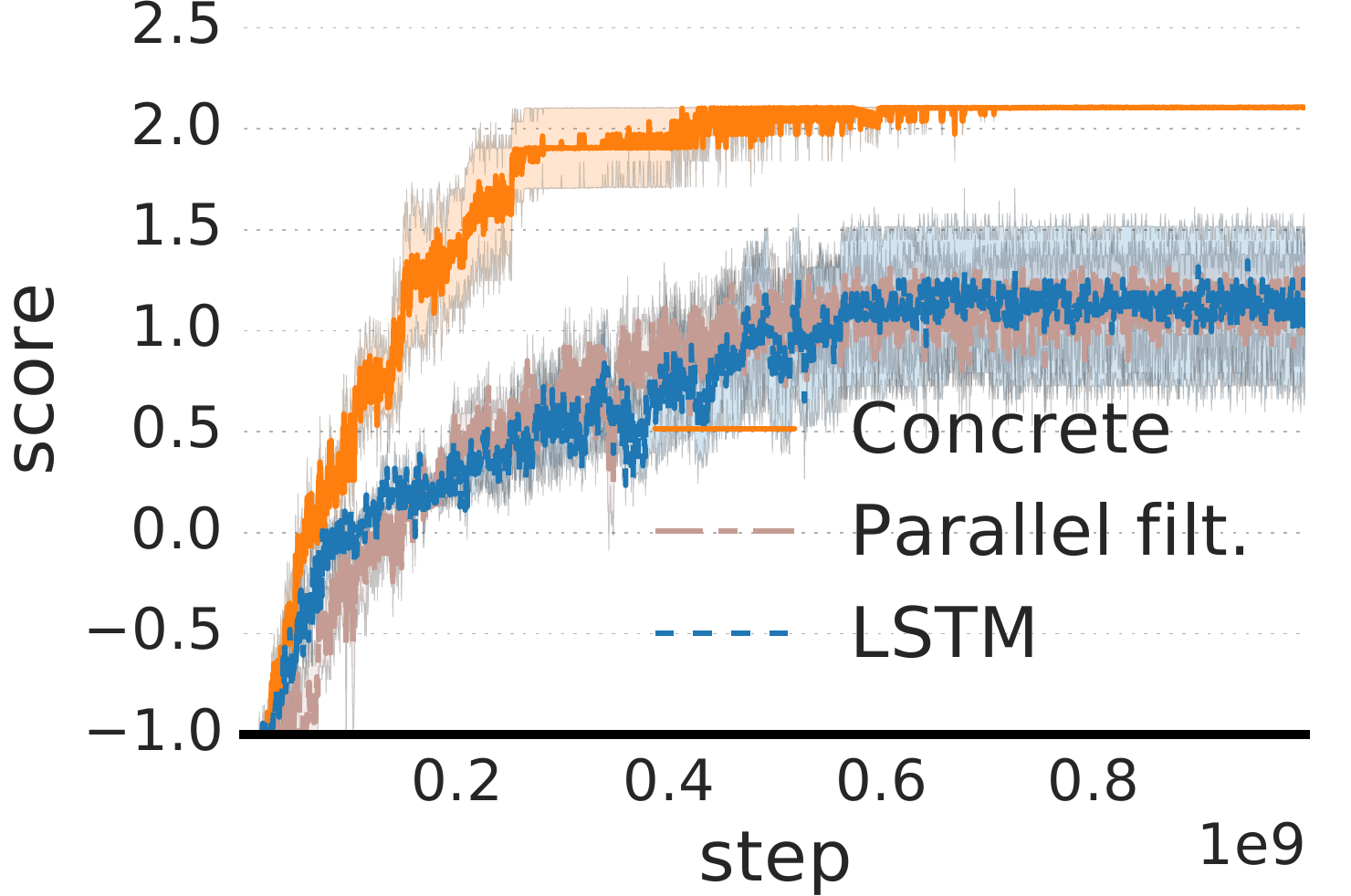}
    \caption*{\em\hspace{4mm}Every 20 steps.}
  \end{subfigure}
  \hspace{0.005\linewidth}
  \begin{subfigure}{0.24\linewidth}
    \includegraphics[width=\linewidth]{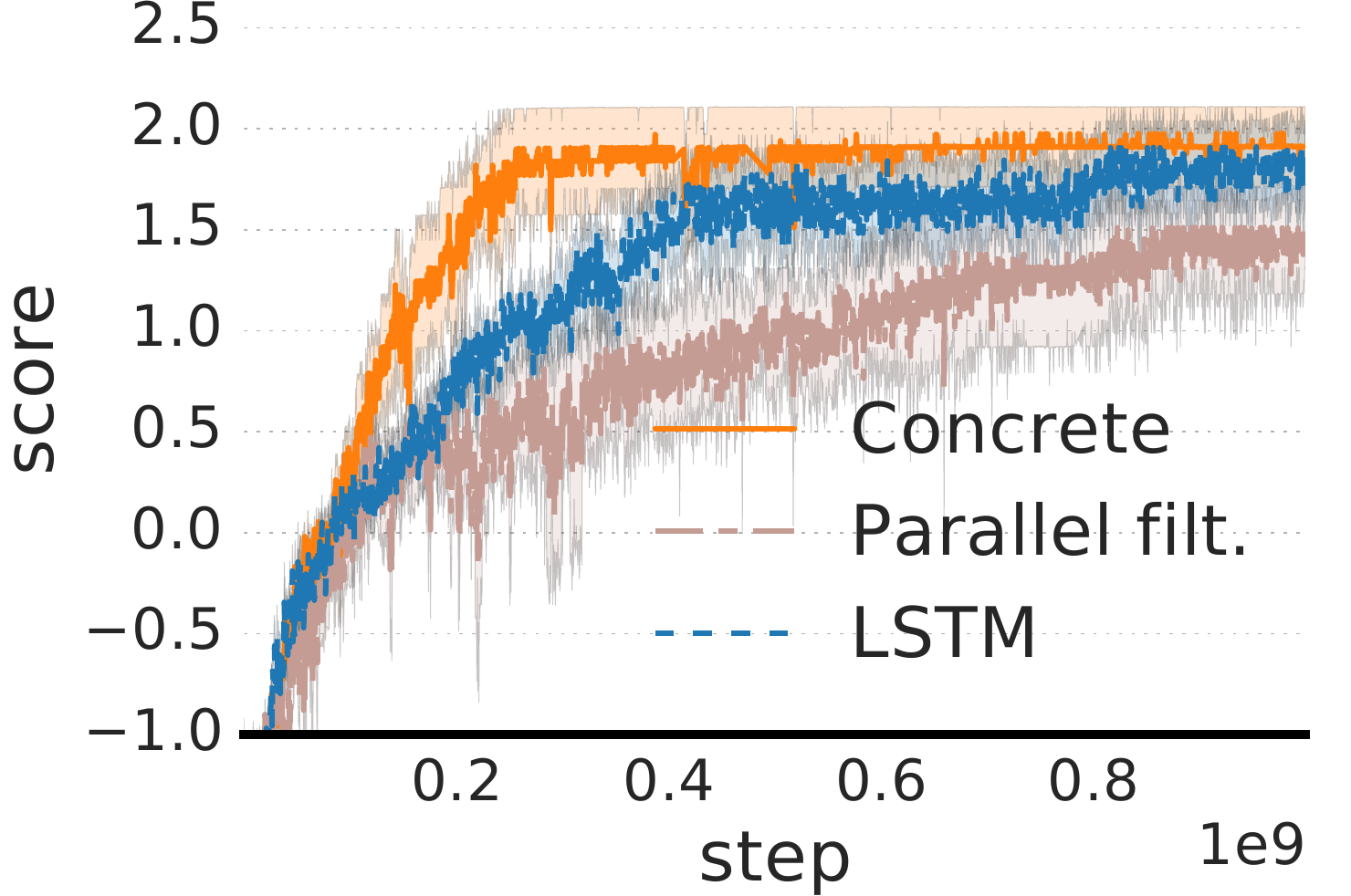}
    \caption*{\em\hspace{4mm}Every 300 steps.}
  \end{subfigure}
\end{figure}

\section{Network schematics}
\label{app:schem}

\begin{figure}[H]
  \centering
  \includegraphics[width=0.35\linewidth]{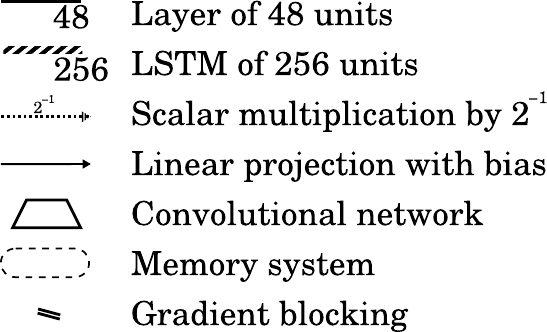}
  \caption{Legend for the symbols used in the neural network schematics in this
           Appendix.}
\end{figure}

\subsection{LP-RNN-based networks used in the experiments}

\begin{figure}[H]
  \centering
  \includegraphics[width=\linewidth]{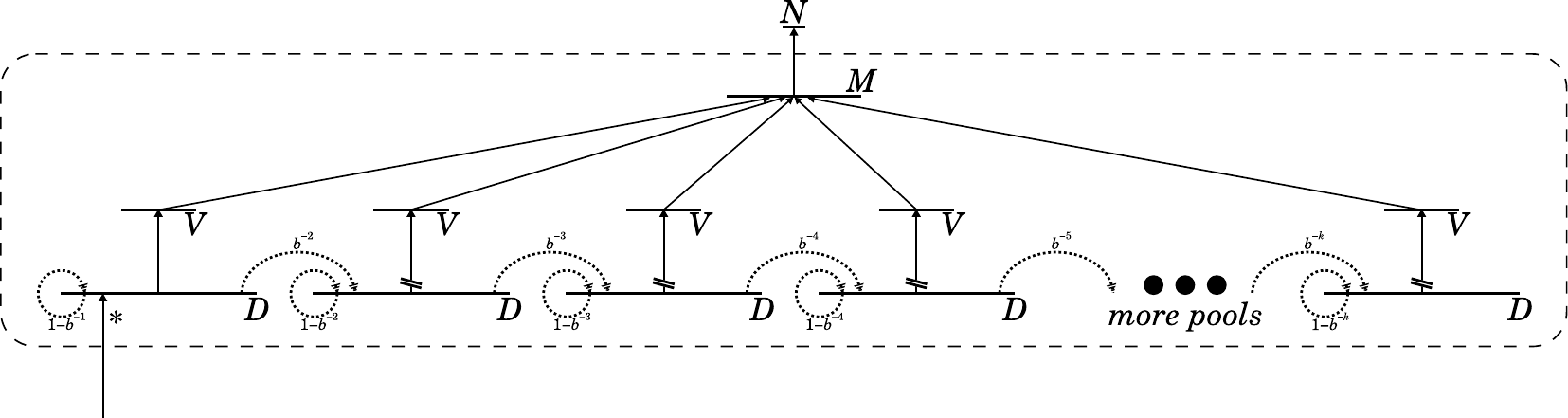}
  \caption{A schematic of the LP-RNN (Concrete) network family applied
           to the sequence classification tasks in Section \ref{sec:expsc}.
           The size of the network output $N$ is determined by the number of
           classes in the classification task; the other parameters shown
           ($M$: summariser size; $V$: viewport size; $D$: pool size;
           $b$: filter base) as well as the number of pools $k$ are varied as
           described in the text. Note that gradients through all but the first
           pool are blocked for computational efficiency, as the low-pass
           architecture will naturally attenuate gradients through most of the
           pools on its own. At the arrow marked *, the one-hot representation
           of the input symbol projects into the first pool via multiplication
           by $b^{-1}I'$ (note no bias), where $I'$ is initialised as a
           ``padded identity'' matrix; however, all of the values in $I'$ may be
           modified by the optimiser.}
  \label{fig:schmidcrete}
\end{figure}

\begin{figure}[H]
  \includegraphics[width=\linewidth]{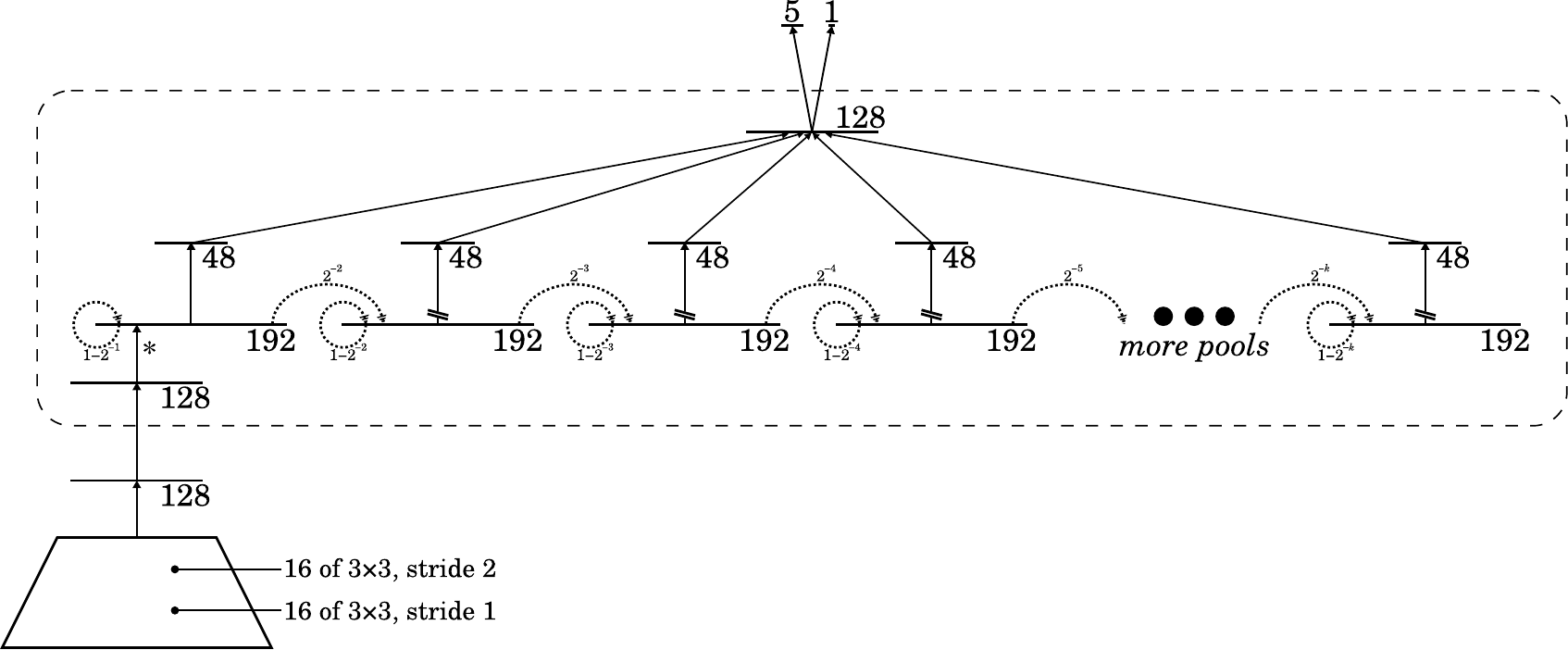}
  \caption{A schematic of the LP-RNN (Concrete) network applied to the
           reinforcement learning tasks in Section \ref{sec:exprl}. The input
           supplied to the convolutional network is a
           rows$\,\times\,$columns$\,\times\,$features binary array. The
           network is an actor-critic network; the five-dimensional output
           comprises unscaled logits for the five agent actions
           ($\uparrow$, $\downarrow$, $\leftarrow$, $\rightarrow$, remain in
           place), while the one-dimensional output is a state value estimate.
           The only adjustable parameter for this architecture is the number of
           pools $k$, which is selected for the task as described in the text.
           Note that gradients through all but the first pool are blocked for
           computational efficiency, as the low-pass architecture will
           naturally attenuate gradients through most of the pools on its own.
           At the arrow marked *, the 128-dimensional embedding of the processed
           input projects into the first pool via multiplication by $2^{-1}I'$
           (note no bias), where $I'$ is initialised as a ``padded identity''
           matrix; however, all of the values in $I'$ may be modified by the
           optimiser.}
  \label{fig:rlcrete}
\end{figure}

\subsection{LSTM-based networks used in the experiments}

\begin{figure}[H]
  \begin{subfigure}{\columnwidth/2}
    \centering
    \vspace{2.6cm}
    \includegraphics[height=2.4cm]{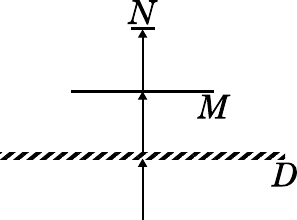}
    \caption{A schematic of the LSTM-based network family applied to the
             sequence classification tasks in Section \ref{sec:expsc}. The size
             of the network output $N$ is determined by the number of classes in
             the classification task; the other parameters shown ($M$: hidden
             layer size; $D$: LSTM size) are varied as described in the text.}
    \label{fig:rllstm}
  \end{subfigure}
  \hspace{2mm}
  \begin{subfigure}{\columnwidth/2}
    \centering
    \includegraphics[height=5cm]{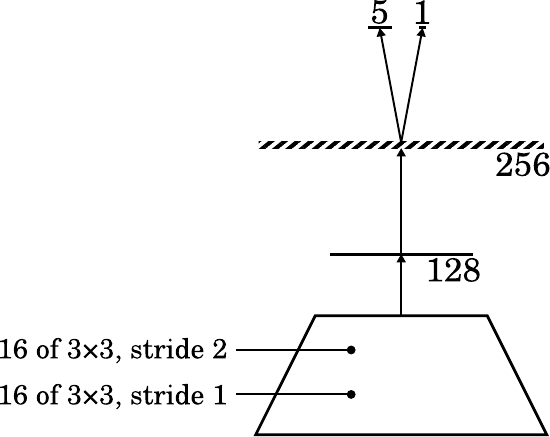}
    \caption{A schematic of the LSTM-based network applied to the reinforcement
             learning tasks in Section \ref{sec:exprl}. This network has the
             same inputs and outputs as the Concrete network in Figure
             \ref{fig:rlcrete}.}
  \end{subfigure}
\end{figure}

\subsection{Parallel filter-based networks used in the experiments}

\begin{figure}[H]
  \centering
  \includegraphics[width=\linewidth]{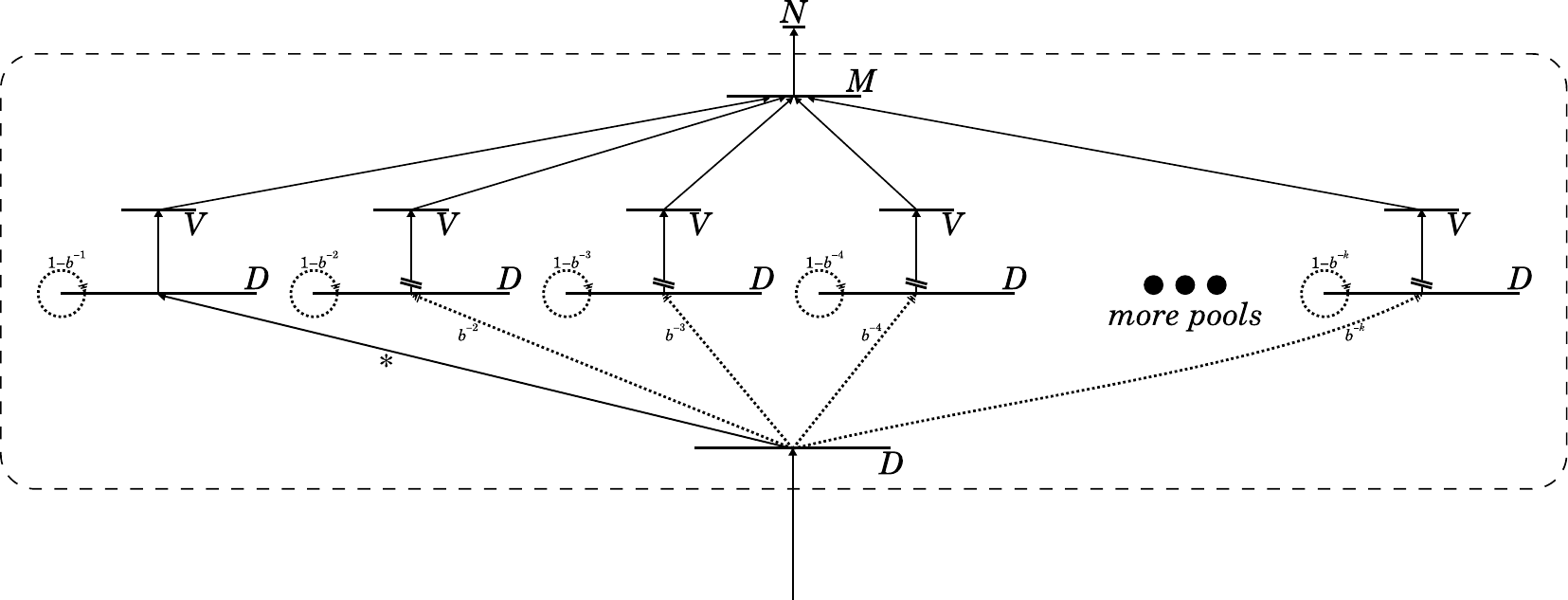}
  \caption{A schematic of the parallel-pool network family applied to the
           sequence classification tasks in Section \ref{sec:expsc}. This
           family has the same structural parameters ($D$, $V$, $k$, $M$) as
           the Concrete network family in Figure \ref{fig:schmidcrete} and
           features an identical pattern of gradient blocking. The layer at
           bottom is a learned $D$-dimensional embedding of the input; its
           projection to the first pool (the arrow marked *) is a learnable
           bias-free projection initialised as $b^{-1}I$, where $I$ is the
           identity matrix.}
\end{figure}

\begin{figure}[H]
  \includegraphics[width=\linewidth]{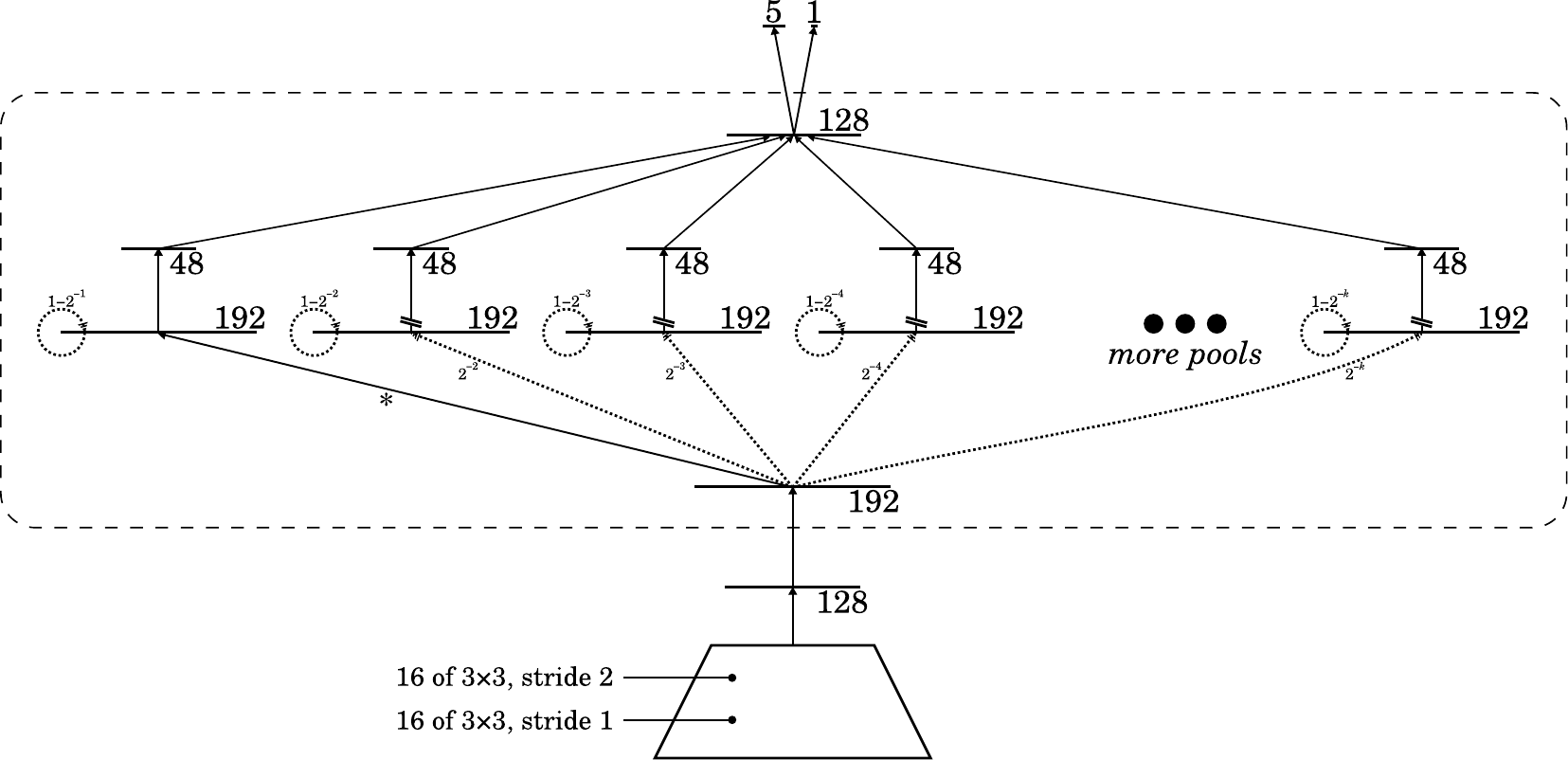}
  \caption{A schematic of the parallel-pool network family applied to the
           reinforcement learning tasks in Section \ref{sec:exprl}. This network
           has the same inputs and outputs as the Concrete network in Figure
           \ref{fig:rlcrete} and features an identical pattern of gradient
           blocking. The layer just upstream of the pools is a 198-dimensional
           embedding of the processed input; its projection to the first pool
           (the arrow marked *) is a learnable bias-free projection initialised
           as $2^{-1}I$, where $I$ is the identity matrix.}
\end{figure}

\section{Population Based Training (PBT)}
\label{app:pbt}

We use an adaptation of PBT to augment the optimisation of network weights for our model (and corresponding baselines) for all RL tasks. 

For each run of the experiment, nine agents are launched in parallel. Agents train for 30,000 episodes each before the evaluation phase. Here, each agent randomly samples another and compares its performance across the last 3,000 episodes -- this is done by performing a t-test over both scores. If the sampled agent is found to be statistically better (p-value $<0.05$), its weights
are copied over. 

Unlike conventional PBT \cite{jaderberg2017pbt}, where agents that copy weights
from better-performing agents also copy (and then perturb) various
hyperparameters as well (e.g. gradient descent step size), our adaptation leaves
agent hyperparameters unchanged. In a sense, this approach realises an
implicit adaptive hyperparameter schedule by allowing well-performing sets of
weights to ``jump'' amongst different hyperparameter configurations.

\section{When LSTM starts working on Cued Catch and T-maze}

Even without blocking gradients through the hidden state at any point within
the 300-step network unrolls described in Section \ref{sec:exprl}, our LSTM
networks showed poor performance on the Cued Catch and T-maze tasks. To
demonstrate that an LSTM has the representational capacity to solve these
tasks, we swept over more relaxed settings of their respective ``difficulty
parameters'' to identify a point at which the LSTM network (with unblocked
300-step unrolls) can succeed.

For Cued Catch, the difficulty parameter is the number of reward-free trials
before the player starts receiving a reward for ``catching'' the correct block.
The original experiment used 40 reward-free trials; in our sweeping, we first
observed some non-random performance at 10 reward-free trials and good
performance at 5:
\begin{figure}[H]
  \begin{subfigure}{0.24\linewidth}
    \includegraphics[width=\linewidth]{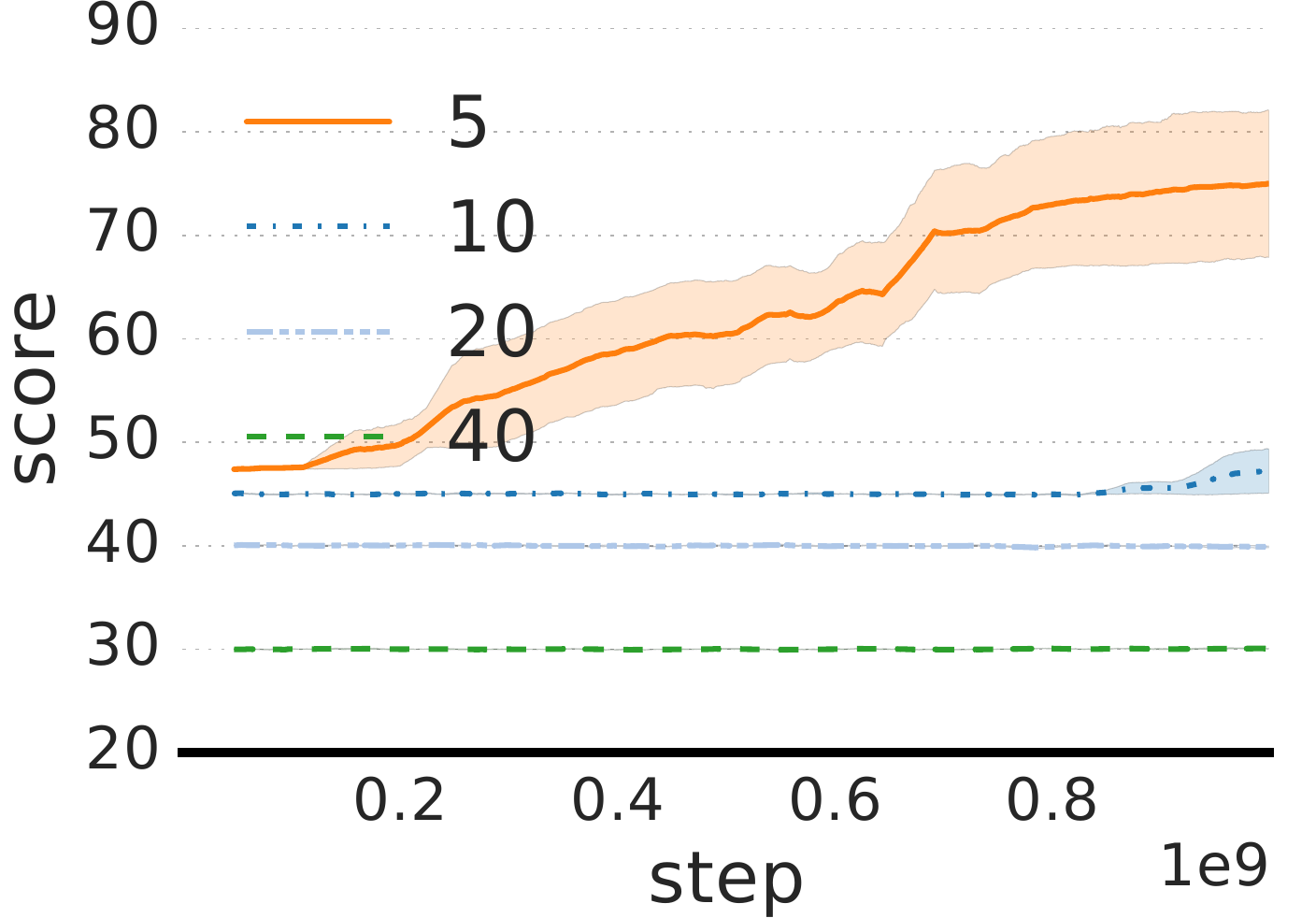}
    \caption*{\em\hspace{4mm}Every 300 steps.}
  \end{subfigure}
\end{figure}
Note that the total number of trials remains the same in all cases, which
accounts for the reason these curves appear to be vertically offset: fewer
reward-free trials means more trials where even a random agent will receive
some reward.

For T-maze, the difficulty parameter is the number of frames the player spends
immobilised in ``limbo''. The original experiment immobilised the agent for
280 frames; we observed gradual performance improvement as this delay was
reduced. By 140 frames, we observed good performance:
\begin{figure}[H]
  \begin{subfigure}{0.24\linewidth}
    \includegraphics[width=\linewidth]{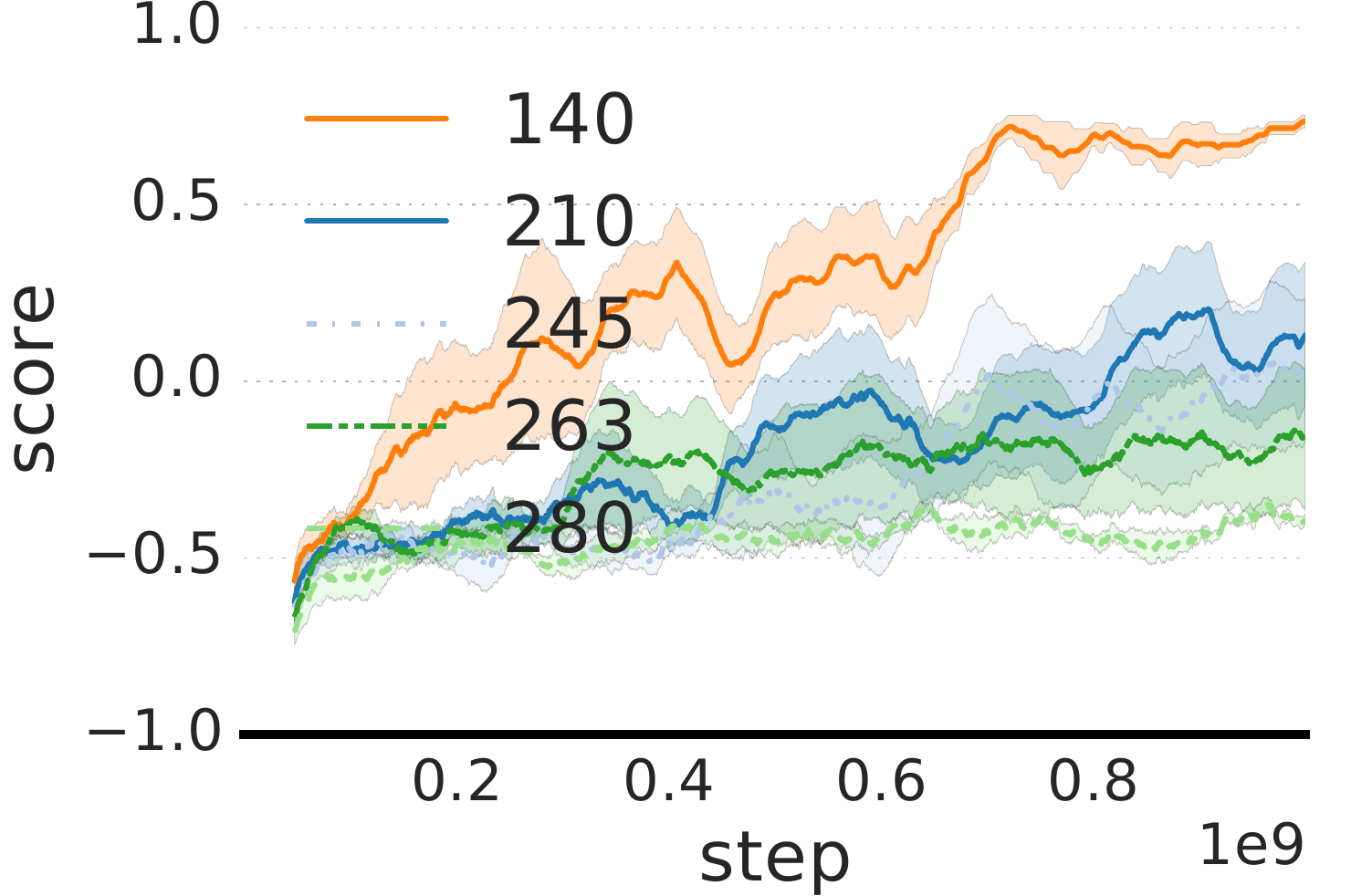}
    \caption*{\em\hspace{4mm}Every 300 steps.}
  \end{subfigure}
\end{figure}

\section{More on LP-RNN as a linear operator}

In all of the following, we consider the input to be a sequence of scalar
floating-point numbers; for vector inputs, we can imagine parallelising the
analysis for each vector dimension. We imagine a memory with no input embedding
and $T$ pools of size 1, where $T$ is the length of the input sequence. Let
$
Y_t = [y_{1,t}, y_{2,t}, y_{3,t}, \ldots y_{T,t}]
$
be the contents of the $T$ pools at timestep $t$. Similarly, if we repackage
each scalar input value $x_t$ into a $T$-vector $X_t$ whose entries are all 0
except for $x_t$ in the first entry, that is
$
X_t = [x_t, 0, 0, 0, \ldots 0],
$
then the memory pool contents at time $t$ can be expressed as the following
recurrence:
\begin{align*}
    Y_0 &= \mathbf{0} \\
    Y_t &= (X_t + Y_{t-1}) \mathbf{M}, \\
\end{align*}
where $\mathbf{M}$ is a matrix describing how values diffuse through the pools
during a single timestep. This matrix is upper-triangular with entries in the
following pattern:
$$
\mathbf{M} =
\left[
\begin{array}{rrrrc}
    (1-b^{-1})
  & b^{-2}(1-b^{-1})
  & b^{-2}b^{-3}(1-b^{-1})
  & b^{-2}b^{-3}b^{-4}(1-b^{-1})
  & \cdots
  \\
  & (1-b^{-2})
  & b^{-3}(1-b^{-2})
  & b^{-3}b^{-4}(1-b^{-2})
  &
  \\
  &
  & (1-b^{-3})
  & b^{-4}(1-b^{-3})
  & \cdots
  \\
  &
  &
  & (1-b^{-4})
  &
  \\
  &
  &
  &
  & \ddots
\end{array}
\right],
$$
or, for indices $i, j$, $\mathbf{M}_{i,j} = (1-b^{-i})\prod_{k=i+1}^j b^{-k}$
if $i \le j$ and 0 otherwise; $b > 1$ is some fixed base. (Recall that the
LP-RNN description in the main text allows the filter coefficients to be
different for each pool; here we limit our
description
to common parameterisations that use powers of a common base.)

Given the linearity of the above recurrence, we can see that
$$
Y_T = X_1\mathbf{M}^T + X_2\mathbf{M}^{T-1} + X_3\mathbf{M}^{T-2} + \ldots +
      X_T\mathbf{M},
$$
and because each $X_t$ is 0 except for the first entry, each term is just
the first row of a power of $\mathbf{M}$ scaled by some integer. Therefore,
if we place the entire input sequence in reverse into some vector
$Z = [x_T, x_{T-1}, x_{T-2}, \ldots x_1]$, we can express $Y_T$ as a single
linear transformation of this $Z$:
$$
Y_T = Z \mathbf{J},\:\:
\mathbf{J} =
\begin{bmatrix}
(\mathbf{M}^{1})_1 \\
(\mathbf{M}^{2})_1 \\
(\mathbf{M}^{3})_1 \\
\vdots
\end{bmatrix}
;
$$
in other words, $\mathbf{J}$ collects in its rows the first rows of the powers
of $\mathbf{M}$.

\end{document}